\documentclass[11pt]{article}

\usepackage{acl}

\usepackage{times}
\usepackage{latexsym}

\usepackage[T1]{fontenc}

\usepackage[utf8]{inputenc}

\usepackage{microtype}

\usepackage{inconsolata}

\usepackage{graphicx}

\usepackage{hyperref}
\usepackage{url}
\usepackage[most]{tcolorbox} 
\usepackage{xcolor}
\usepackage{xspace}
\usepackage{longtable} 
\usepackage{booktabs}  
\usepackage{multirow}  
\usepackage{array}     
\usepackage{float} 
\usepackage{ragged2e}
\usepackage{enumitem}
\usepackage{xcolor}
\usepackage{tabularray}
\UseTblrLibrary{booktabs}
\usepackage{makecell}
\usepackage{cleveref}

\usepackage{placeins} 

\newcommand{\dsetname}{\textsc{Updesh}\xspace}
\newcommand{\phiBM}{\textsc{Phi4-14B}\xspace}
\newcommand{\llamaBM}{\textsc{Llama-3.1-8B}\xspace}
\newcommand{\llamaIXL}{\textsc{Llama-3.1-405B-Instruct}\xspace}
\newcommand{\llamaIL}{\textsc{Llama-3.3-70B-Instruct}\xspace}

\newcommand{\gpto}{\textsc{GPT-4o}\xspace}

\newcommand{\qwenIXL}{\textsc{Qwen3-235B-A22B}\xspace}
\newcommand{\orcaagent}{\textsc{OrcaAgent-Instruct}\xspace}
\newcommand{\lima}{\textsc{LIMA-X}\xspace}
\newcommand{\orcamath}{\textsc{OrcaMath}\xspace}
\newcommand{\aya}{\textsc{Aya-Collection}\xspace}
\newcommand{\bactrian}{\textsc{Bactrian-X}\xspace}
\newcommand{\malpaca}{\textsc{mAlpaca}\xspace}
\newcommand{\alpaca}{\textsc{Alpaca}\xspace}
\newcommand{\indicalign}{\textsc{IndicAlign}\xspace}
\newcommand{\indictrans}{\textsc{IndicTrans2}\xspace}

\title{\textbf{UPDESH}: Synthesizing Grounded Instruction Tuning Data for 13 Indic Languages}

\author{
  \parbox{0.9\linewidth}{\centering 
    Pranjal A. Chitale$^{\spadesuit}$ \quad 
    Varun Gumma$^{\heartsuit}$\thanks{Work done at Microsoft} \quad 
    Sanchit Ahuja$^{\diamondsuit}$\footnotemark[1] \quad 
    Prashant Kodali$^{\spadesuit}$ \\[0.3em]
    Manan Uppadhyay$^{\spadesuit}$ \quad 
    Deepthi Sudharsan$^{\clubsuit}$\footnotemark[1] \quad 
    Sunayana Sitaram$^{\spadesuit}$\thanks{Corresponding author} \\[0.5em]
    $^{\spadesuit}${\rm Microsoft Corporation} \quad 
    $^{\heartsuit}${\rm Nanyang Technological University} \quad 
    $^{\diamondsuit}${\rm Northeastern University} \quad 
    $^{\clubsuit}${\rm Independent Researcher} \\
    {\tt \small pranjalchitale@gmail.com, sunayana.sitaram@microsoft.com}
  }
}

\begin{document}
\maketitle

\begin{abstract}
Developing culturally grounded multilingual AI systems remains challenging, particularly for low-resource languages. While synthetic data offers promise, its effectiveness in multilingual and multicultural contexts is underexplored. We investigate bottom-up synthetic data generation using large open-source LLMs ($\geq 235$B parameters) grounded in language-specific Wikipedia content, complementing dominant top-down translation-based approaches from English. We introduce $\dsetname$, a high-quality large-scale synthetic instruction-following dataset comprising 9.5M data points across 13 Indian languages and English, encompassing diverse reasoning and generative tasks. Comprehensive evaluation using automated metrics and 10K human assessments confirms high data quality. Downstream evaluations performed by fine-tuning models on various datasets and assessing performance across 13 diverse multilingual datasets and model comparative evaluations, demonstrate that models trained on $\dsetname$ consistently obtain significant improvements on NLU, NLG evaluations. Finally, through ablation studies and cultural evaluations, we show that context-aware, culturally grounded data generation is essential for effective multilingual AI development .
\end{abstract}
\section{Introduction}
Building multilingual, multicultural AI is essential for equitable access across communities. Yet frontier models often underperform in non-English and non-Western settings because diversity is limited in pre-training corpora and English-centric choices pervade the development pipeline \citep{joshi-etal-2020-state}. While large-scale crawling can expand pre-training data, fine-tuning and evaluation sets require deliberate construction; translation-only approaches often overlook linguistic nuance and cultural context.

\citet{joshi-etal-2020-state} identify a stark imbalance in web-scale pretraining data: their lowest-resource categories (Classes 5–6) cover over 2.4K languages (93.87\% of the world’s languages) spoken by 1.2B people, yet remain severely underrepresented online. This gap is more pronounced for fine-tuning and evaluation datasets \citep{hu2025quantifyinglanguagedisparitiesmultilingual}. Synthetic data shows promise in English for reasoning \citep{goldie2025syntheticdatageneration,harsha-etal-2025-synthetic}, coding \citep{wei2024magicoder,shao-etal-2025-case2code}, and retrieval \citep{bonifacio2022inparsdataaugmentationinformation,dai2023promptagator,chitale2025evaluatingeffectivenessscalabilityllmbased}, but pipelines often embed English-centric quality assumptions that may not transfer. Evaluating multilingual, multicultural synthetic data thus remains open: standard human and automatic checks (fluency, correctness, diversity) are often insufficient, and downstream utility must be validated via fine-tuning and benchmarking. 

\begin{figure*}
    \centering
    \includegraphics[width=1\linewidth]{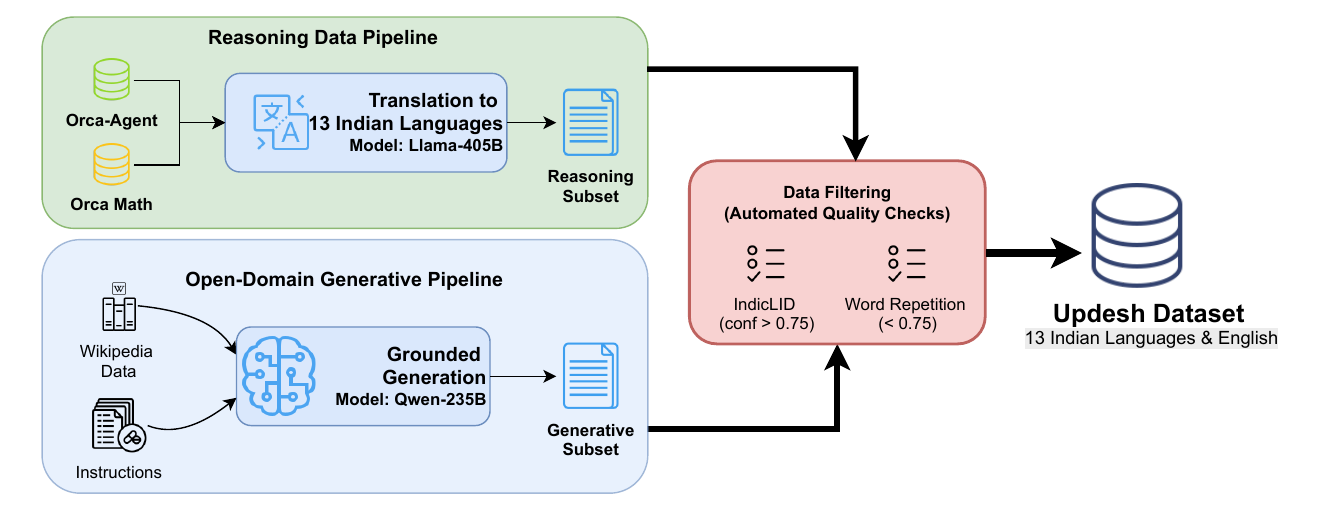}
    \caption{Overview of the data generation pipeline for the Updesh Dataset.}
    \label{fig:dataset_overview}
\end{figure*}

In this work, we introduce \dsetname, a culturally grounded multilingual synthetic instruction dataset with 9.5M examples, and provide extensive analyses of its quality and downstream utility. Models fine-tuned on \dsetname consistently improve NLG and NLU performance, across languages and shows a considerable uplift in cultural evaluations as well. Alongside the dataset, we distill a set of design considerations for multilingual and multicultural synthetic data generation spanning generation strategies, language-specific grounding, quality assessment, and evaluation. Our results also expose limitations of LLM-as-judge in culturally nuanced settings and yield actionable guidance for dataset design, filtering, and downstream validation, enabling reproducible alternatives to translation-first pipelines.

\section{Related Work}
\label{sec:related_works}

\paragraph{English Instruction Fine-tuning (IFT).}
IFT adapts pre-trained LMs to follow instructions using instruction–response pairs \citep{ouyang2022traininglanguagemodelsfollow}. Early English IFT corpora include FLAN \citep{wei2022finetuned}, which scaled to $>$1.8k tasks with CoT prompting, and Self-Instruct \citep{wang-etal-2023-self-instruct}, which showed the viability of LLM-generated synthetic data and inspired Stanford \alpaca \citep{alpaca} and Alpaca-GPT4 \citep{peng2023instruction}. Follow-on work explored conversational and curation-heavy routes: Vicuna \citep{vicuna2023}, WizardLM’s Evol-Instruct \citep{xu2024wizardlm}, LIMA \citep{zhou2023lima} showing that $\sim$1k high-quality prompts suffice, and the ORCA series \citep{mukherjee2023orca,mitra2023orca2}, which introduced explanation tuning and prompt erasure, culminating in $\orcaagent$ with 25.8M synthetic pairs \citep{mitra2024agentinstructgenerativeteachingagentic}.

\paragraph{Multilingual IFT Datasets.}
Multilingual instruction-following has been pursued via (i) \textit{translation}, (ii) \textit{template/synthesis}, and (iii) \textit{hybrids}. Translation-focused efforts include $\bactrian$ \citep{li2023bactrianx}, which translates \alpaca \citep{alpaca} and Dolly \citep{DatabricksBlog2023DollyV2} into 3.4M pairs over 52 languages, and \malpaca \citep{chen-etal-2024-monolingual}, which translates \alpaca. Template-driven generation such as M2Lingual extends Evol-guided taxonomies \citep{xu2024wizardlm} to 70 languages. Hybrid pipelines combine crowdsourcing, templating, and translation: $\aya$ \citep{singh-etal-2024-aya} integrates crowd data across 65 languages with repurposed xP3 \citep{muennighoff-etal-2023-crosslingual}, FLAN \citep{flan}, and Dolly, using NLLB 3.3B \citep{nllb} for N-way translation; $\indicalign$ aggregates 74.7M prompt–response pairs for 20 Indian languages via dataset aggregation, \indictrans{}-based translation \citep{gala2023indictrans}, synthetic conversations from India-centric Wikipedia, and crowdsourcing.

\paragraph{Data Generation Strategies.}
Most prior work distills outputs from stronger teachers (e.g., \textsc{GPT-4}). Recent alternatives mitigate distillation limits by leveraging diverse web content with self-augmentation and self-curation: Instruction Backtranslation \citep{li2024selfalignment} synthesizes instructions from documents, and Back-and-Forth Translation \citep{nguyen-etal-2024-better} iteratively rewrites responses with LLMs, often outperforming pure distillation.

\paragraph{Limitations of Prior Multilingual IFT.}
Translation-heavy datasets (e.g., \bactrian, \malpaca, and even curation-focused \lima) tend to emphasize basic instruction following, underrepresent advanced reasoning, and provide limited demographic/cultural grounding. Sentence-level MT (e.g., NLLB 3.3B, \indictrans) can introduce context-loss and subtle errors that propagate during training. Despite broad coverage, $\aya$ contains comparatively little culturally specific content, while $\indicalign$ relies heavily on WordNet \citep{miller-1994-wordnet} and QA-style prompts, limiting task diversity. Finally, most corpora are short-context and single-turn, leaving long-context and multi-turn underexplored.

\begin{table*}[!htbp]
\centering
\small
\begin{tabular}{lrrr|lrrr}
\toprule
\multicolumn{4}{c|}{\textbf{Reasoning Subset (13 Indian Language \& English)}} &
\multicolumn{4}{c}{\textbf{Generative Subset (13 Indian Language \& English)}} \\
\midrule
\textbf{Category} & \textbf{Total} & \textbf{Drop (\%)} & \textbf{Final} &
\textbf{Category} & \textbf{Total} & \textbf{Drop (\%)} & \textbf{Final} \\
\midrule
\textsc{Analytical R} & 350K & 0.047 & 349.8K &
\textsc{Logical R} & 229.4K & 1.459 & 226.0K \\

\textsc{Brain Teaser} & 700K & 0.043 & 699.7K &
\textsc{Multihop QA} & 229.4K & 1.459 & 226.0K \\

\textsc{Fermi} & 350K & 0.015 & 349.9K &
\textsc{Creative Writing} & 229.4K & 1.459 & 226.0K \\

\textsc{Fs-CoT-Flow} & 350K & 3.769 & 336.8K &
\textsc{Multi-turn Dialogue} & 229.4K & 1.611 & 225.7K \\

\textsc{Math} & 2.80M & 0.035 & 2.80M &
\textsc{Summarization} & 229.4K & 1.526 & 225.9K \\

\textsc{MCQ} & 1.40M & 0.135 & 1.40M &
\textsc{Translation (to En)} & 229.4K & 0.641 & 227.9K \\

\textsc{Reading Comp.} & 700K & 0.379 &   &
\textsc{Translation (from En)} & 229.4K & 17.047 & 190.3K \\

\textsc{Text Classification} & 700K & 1.878 & 686.9K &
\textsc{Cultural MHR} & 375.7K & 0.347 & 374.4K \\

& & & &
\textsc{Causal R} & 229.4K & 1.453 & 226.0K \\

\midrule
\textbf{Total} & \textbf{7.35M} & -- & \textbf{7.32M} &
\textbf{Total} & \textbf{2.21M} & -- & \textbf{2.15M} \\
\bottomrule
\end{tabular}
\caption{Document filtering statistics aggregated over all 13 Indic languages and English.
Totals and final counts are reported using K (thousands) and M (millions) notation.
Per-language counts are uniform within each category and described in the text.
MHR denotes Multi-Hop Reasoning, R denotes Reasoning.}
\label{tab:filtering_stats_combined}
\end{table*}

\section{Data Generation}
\label{sec:data_gen}
We synthesize $\dsetname$, a dataset covering 13 Indic languages—Assamese, Bengali, Gujarati, Hindi, Kannada, Malayalam, Marathi, Nepali, Odia, Punjabi, Tamil, Telugu, and Urdu. For each language, $\dsetname$ includes two complementary subsets targeting distinct facets of multilingual instruction following: \textit{reasoning} and \textit{open-domain generation}. This design recognizes that reasoning capabilities are largely language- and culture-agnostic, making translation-based approaches suitable for tasks like mathematical problem-solving and logical inference \citep{shaham-etal-2024-multilingual}. We summarize the key design considerations guiding the curation of $\dsetname$ in Appendix~\ref{appendix:design_considerations} along with Figure~\ref{fig:framework}.

Existing high-quality reasoning datasets such as $\orcaagent$ and $\orcamath$, thus are valuable resources for multilingual adaptation.
However, generative capabilities requiring cultural awareness, linguistic naturalness, and factual grounding in local contexts cannot be adequately addressed through translation due to inherent Western-centric biases and lack of cultural specificity in existing datasets \citep{yao-etal-2024-benchmarking}. Therefore, our generative subset employs a grounded approach that ensures factuality through Wikipedia content, maintaining linguistic naturalness through native language generation, and preserves cultural adherence through systematic curation of India-specific cultural artifacts.



\paragraph{Reasoning Data} Inspired by prior work \citep{ahuja-etal-2025-sphinx, khan-etal-2024-indicllmsuite}, we translate eight subsets of the $\orcaagent$ \citep{mitra2024agentinstructgenerativeteachingagentic} and $\orcamath$ \citep{mitra2024orcamathunlockingpotentialslms} datasets into 13 Indic languages. Specifically, we consider seven reasoning-related subsets from $\orcaagent$\footnote{\url{https://huggingface.co/datasets/microsoft/orca-agentinstruct-1M-v1}} along with the Math subset from $\orcamath$\footnote{\url{https://huggingface.co/datasets/microsoft/orca-math-word-problems-200k}} (\Cref{tab:reasoning_tasks}). Both datasets have been attributed to induce significant chain-of-thought and reasoning capabilities in models during instruction-tuning without the need for specific preference optimization. We employ $\llamaIXL$ for selective translation given its strong coverage in Indian languages and instruction-following capabilities that enable adaptation to various conversational styles \citep{sankar-etal-2025-towards}. Post generation, all outputs undergo strict quality checks, as described in detail below. 





\begin{figure*}[!htbp]
    \centering
    \includegraphics[width=0.8\linewidth]{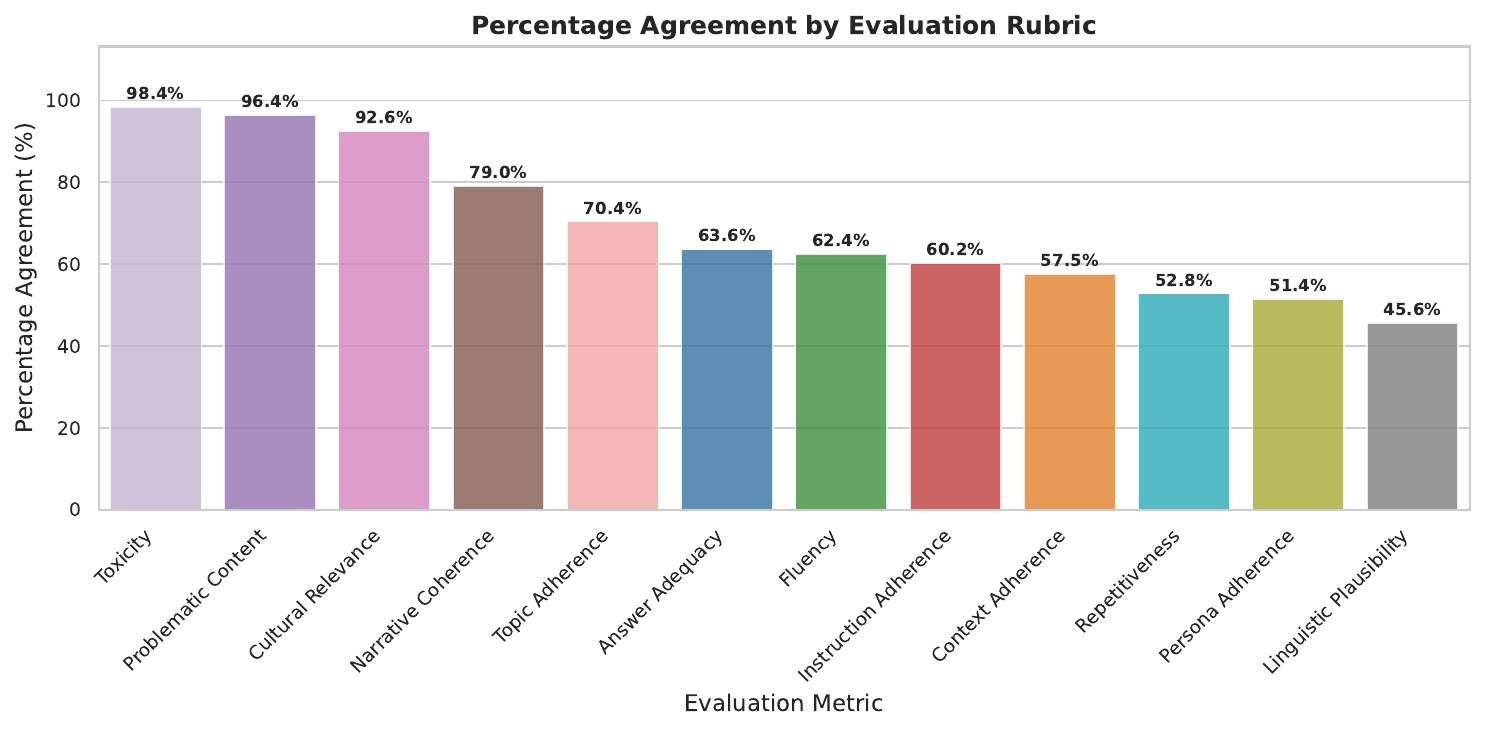}
    \caption{Human LLM-judge agreement across evaluation metrics, revealing differences across dimensions.}
    \label{fig:metric_agreement}
\end{figure*}

\paragraph{Open-Domain Generative Data} Synthesizing generative data poses greater challenges than translation due to increased risks of hallucinations, factual inaccuracies, and demographic misalignment. We compared $\llamaIXL$ and $\qwenIXL$ across reasoning and non-reasoning paradigms, finding $\qwenIXL$ superior for generative tasks and complex instruction following due to stronger reasoning traces. (Also supported by \cite{10.5555/3692070.3692401}).

Inspired by instruction backtranslation techniques \citep{li2024selfalignment}, we construct questions from unlabelled content followed by LLM-generated answers. To ensure diversity, contextual grounding, factual accuracy, and demographic relevance, we leverage Wikipedia pages in target languages as our knowledge base. Table \ref{tab:generative_tasks} summarizes eight generative task categories, with some requiring two LLM inference phases. Further, to ensure cultural representation, we systematically curated culturally relevant content from Wikipedia using the MediaWiki API. 

Following \citep{yao-etal-2024-benchmarking}'s cultural taxonomy, we traversed Wikipedia categories from \texttt{Category:Culture of India} and \texttt{Category:Culture of India by state or union territory}, exploring 2-3 levels deep. This yielded diverse region-specific content spanning festivals, cuisine, traditional arts, architecture, and religious practices. We sampled 26.8K cultural artifacts to create multi-hop question-answer pairs for synthetic data generation. For English, we reuse the reasoning data from $\orcaagent$ \citep{mitra2024agentinstructgenerativeteachingagentic} and $\orcamath$ \citep{mitra2024orcamathunlockingpotentialslms} as-is, while generating the English generative subset from scratch using the same pipeline as for the Indian languages.
\paragraph{Data Filtering}
After generating data points at scale across 13 languages for both the Reasoning and Open-Domain subsets, manual validation was not feasible, therefore, following the approach of \citet{shen2025dcad2000multilingualdataset2000}, we employed automated quality checks but use the standard threshold-based method instead of their anomaly detection-based method. Specifically, we applied two filtering criteria: (1) Language Identification using \textsc{IndicLID} \citep{madhani-etal-2023-bhasa} with a 0.75 confidence threshold, and (2) word repetition ratio capped at 0.75 to flag low-quality generations. 

Table \ref{tab:filtering_stats_combined} shows the filtering results, demonstrating high data quality with drop rates below 2\% for most subsets. The main exception is the \textsc{FS-CoT} subset for Urdu, where the outputs showed excessive repetition leading to higher filtering rates, but we maintain these thresholds to ensure data integrity. For the English-to-XX translation tasks, Assamese had the highest drop rate as the model frequently generated Bengali text instead, likely due to the shared script and similarity between these languages, and because Assamese is a low-resource language.

\section{Dataset Quality Analysis (Q-A)}

\subsection{Q-A for Reasoning Data}
For the reasoning subset, we performed large-scale \textit{selective translation} using $\llamaIXL$. Given inputs with long contexts and non-standard text, we rigorously evaluated translation quality through backtranslation. We randomly selected 4,096 samples per subset and language, backtranslated them to English using $\llamaIL$ (chosen for faster inference and conservative quality bounds), and compared with original sources. Translation fidelity was measured using ChrF~\citep{popovic-2015-chrf} via SACREBLEU~\citep{post-2018-call}. Table~\ref{tab:qc_backtranslation_reasoning} shows consistently high backtranslation scores across all languages, confirming robust translation quality.


\subsection{Q-A for Generative Data}
While large language models (LLMs) have become scalable evaluators under the LLM-as-a-judge paradigm, their reliability in culturally nuanced and low-resource settings remains limited~\citep{watts-etal-2024-pariksha, whitehouse2025menlopreferencesproficiency}. We therefore combined LLM evaluation with native-speaker annotation and measured inter-annotator agreement. Using stratified sampling, we drew 100 instances per category—\textsc{Creative Writing}, \textsc{Cultural Multi-Hop Reasoning}, \textsc{Multi-Turn Dialogue}, and aggregated reasoning (\textsc{Logical}, \textsc{Causal}, \textsc{MultiHop})—across five languages (Assamese, Gujarati, Hindi, Malayalam, Punjabi), yielding 2K items.
Sampling preserved response-length distributions via quintile bucketing. Native-speaker evaluations were conducted via an external agency (Table~\ref{tab:participant-compact}), while \gpto\ served as an automated evaluator using identical protocols for comparability. We defined task-specific, multidimensional metrics using a 3-point Likert scale (0–2) with consistent rubrics (Section~\ref{tab:rubrics}), ensuring a thorough quality check of the generated data. Prompts and annotation guidelines are provided in Appendix~\ref{sec:sample-prompts} and supplementary material. Across 10K individual ratings, only 27 received a zero (0.27\%), indicating uniformly high data quality. We will publicly release all evaluation data—including both human and GPT assessments—to promote research on calibrating LLM-based evaluators.


\textbf{Inter-Annotator Agreement}
To assess the reliability of automated evaluation, we computed percentage agreement between the LLM-judge and human evaluators. Agreement varies notably across metrics (Figure~\ref{fig:metric_agreement}), declining for culturally and linguistically nuanced aspects such as linguistic plausibility and repetition detection in long dialogue sequences. In contrast, objective criteria like toxicity detection and problematic content identification show strong alignment. As the data is derived from benign prompts and topics, toxic or problematic instances are expected to be rare or absent. These findings align with prior work on the difficulty of subjective versus objective evaluation and highlight persistent limitations of current LLM-judges in assessing culturally sensitive content~\citep{watts-etal-2024-pariksha}. The complete distribution of scores across tasks and languages for both evaluations is provided in Appendix~\ref{sec:quality_eval}.

\begin{figure*}[t]
    \centering
    \includegraphics[width=1.0\linewidth]{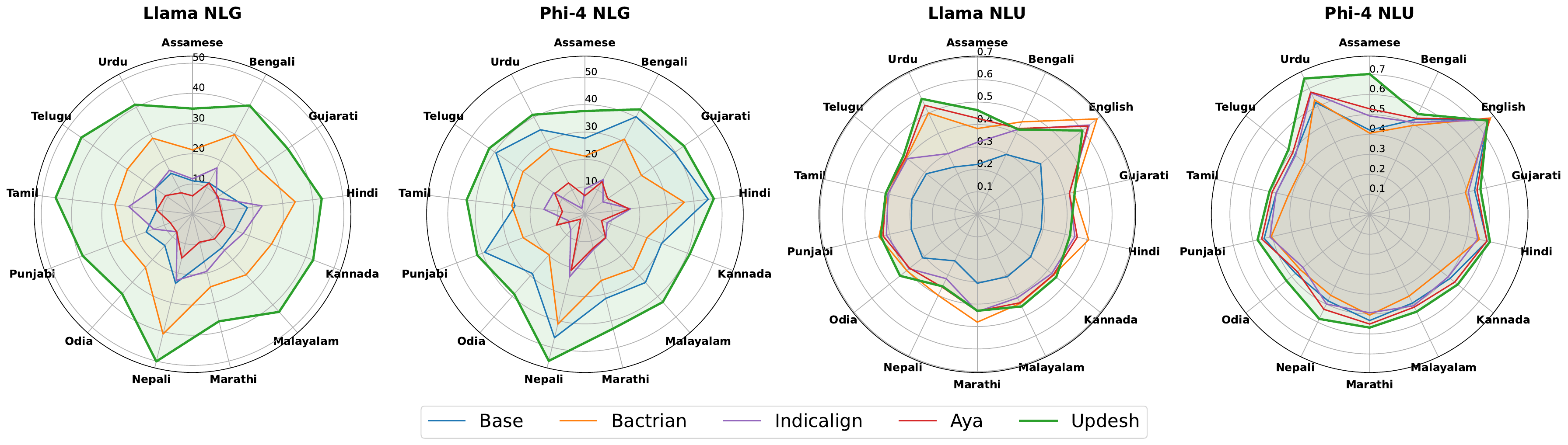}
    \caption{Evaluation plots for models finetuned on \dsetname vs existing datasets}
    \label{fig:model_performance_radar}
\end{figure*}

\begin{figure*}[t]
    \centering
    \includegraphics[width=\textwidth]{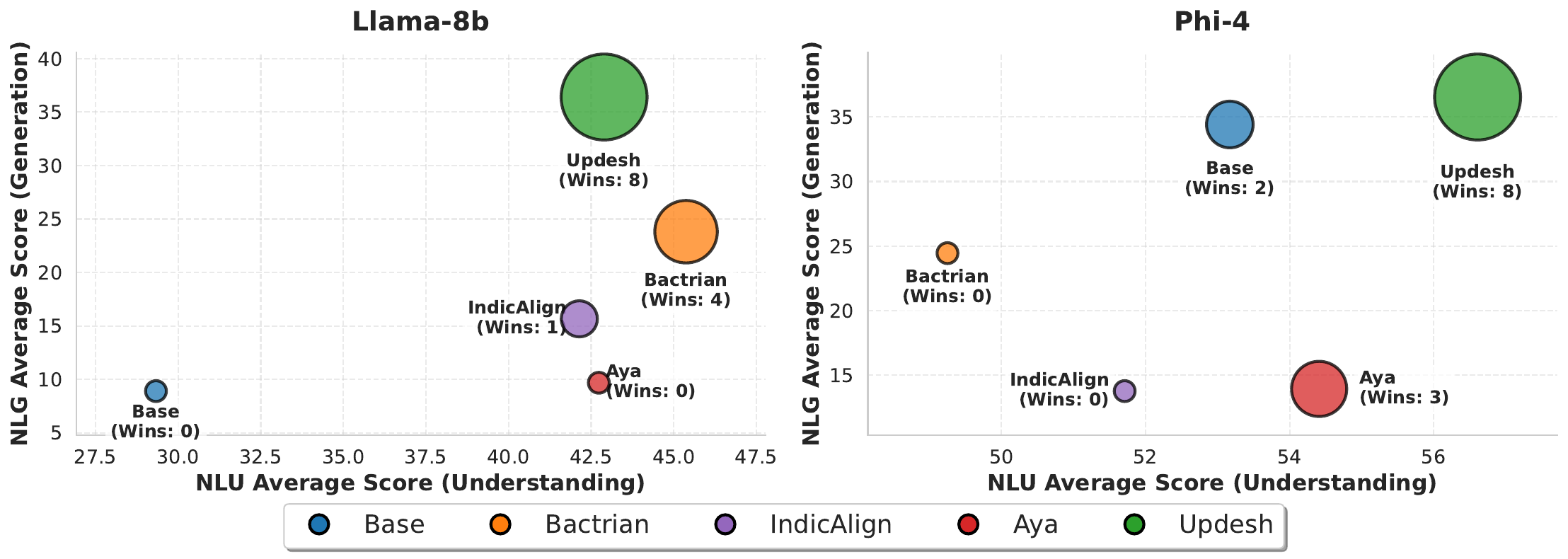}
    \caption{\textbf{Model Performance Landscape: NLU vs. NLG vs. Win Counts.} 
    The horizontal axis represents the average NLU  (accuracy between 0-100), while the vertical axis represents the average NLG score (ChrF between 0-100). The size of each bubble corresponds to the number of specific datasets (12 tasks evaluated) where that the model outperformed all others. \textbf{\dsetname} model (green) demonstrates the most dominant position with high generation scores and the largest number of task wins across both Llama and Phi settings.}
    \label{fig:model_performance_scatter}
\end{figure*}

\begin{figure}[h]
    \centering
    \includegraphics[width=\linewidth]{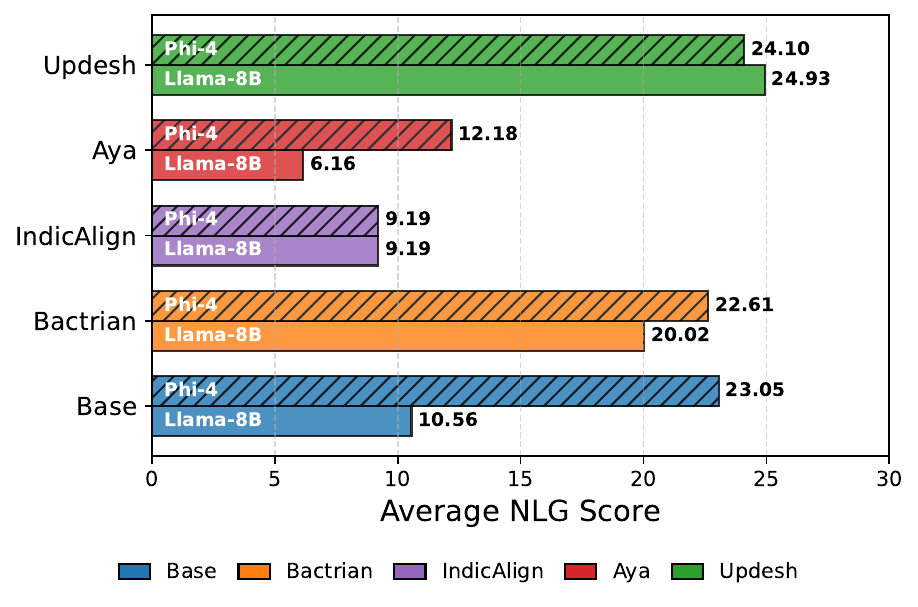}
    \caption{\textbf{NLG performance across 16 out-of-domain Indic languages on Flores.} \dsetname (red) achieves the highest average scores on both Llama-3-8B and Phi-4 architectures, outperforming standard baselines (Zero-shot) and comparable instruction-tuned models (Bactrian, Aya, IndicAlign).}
    \label{fig:updesh_comparison}
\end{figure}

\section{Downstream Tasks Evaluation}
\label{sec:model_training}
We selected two base models, $\llamaBM$ and $\phiBM$, for fine-tuning experiments. These models were chosen based on their size (for feasibility of experiments given available resources) and\textbf{\textit{ }}reported multilingual capabilities \citep{grattafiori2024llama3herdmodels, abdin2024phi4technicalreport}. We used the \texttt{Axolotl} framework\footnote{\url{https://github.com/axolotl-ai-cloud/axolotl}} for all fine-tuning runs. Details regarding the hyperparameters and compute resources used can be found in Appendix \ref{app:hyperparameters}.

\paragraph{Baselines} We fine-tuned both $\llamaBM$ and $\phiBM$ on three high-quality, open-source IFT datasets: $\aya$ \citep{singh-etal-2024-aya}, $\indicalign$ \citep{khan-etal-2024-indicllmsuite} and $\bactrian$ \citep{li2023bactrianx}. To the best of our knowledge, these are the only open datasets that offer both broad language coverage and an instruction-following format.
$\bactrian$ covers 10 of our 13 languages (excluding Assamese, Kannada, Punjabi), while $\aya$ includes all except Punjabi. Since $\aya$ contains millions of samples per language, we uniformly sub-sampled it to 7M samples to create a balanced dataset comparable to $\dsetname$. Similarly, for $\indicalign$, since the WordNet subset ($\sim$97M pairs) is disproportionately large, less diverse, and redundant, we downsampled it to one instance per entry, yielding 7.3M training pairs when combined with its remaining subsets.

\paragraph{Downstream Tasks} Our evaluation framework consists of three task categories to comprehensively assess model capabilities. Natural language understanding (NLU) tasks use multiple-choice questions to measure comprehension and reasoning through likelihood-based scoring. Natural language generation (NLG) tasks, such as translation and summarization, assess models' ability to generate coherent and contextually appropriate outputs. We augment standard dataset-NLU and NLG evaluations with comparative evaluations to understand model win rates. 
This design identifies model strengths and weaknesses across diverse tasks, providing a holistic performance assessment. Dataset details are in Table \ref{tab:eval_datasets}. 

\begin{table}[h]
\centering
\tiny
\resizebox{\columnwidth}{!}{%
\begin{tabular}{lll}
\toprule
& \textbf{Dataset} & \textbf{Source} \\
\midrule
\multirow{7}{*}{\textbf{NLU}}
& MMLU Indic (MMLU-I) & \citet{huggingfaceIndicevalsSarvamai} \\
& ARC Indic (ARC-I) & \citet{huggingfaceIndicevalsSarvamai} \\
& BoolQ Indic (BoolQ-I) & \citet{huggingfaceIndicevalsSarvamai} \\
& TriviaQA Indic (TVQA-I) & \citet{huggingfaceIndicevalsSarvamai} \\
& BeleBele (Bele) & \citet{bandarkar-etal-2024-belebele} \\
& INCLUDE (INCL) & \citet{romanou2025include} \\
& Global MMLU (GMMLU) & \citet{singh-etal-2025-global} \\
\midrule
\multirow{4}{*}{\textbf{NLG}}
& Extreme Summarization (Xsum) & \citet{singh-etal-2024-indicgenbench} \\
& Flores English to Others (Flores EnXX) & \multirow{2}{*}{\citet{goyal-etal-2022-flores}} \\
& Flores Others to English (Flores XXEn) &  \\
& IN22-Conv (IN22-Conv-Doc) - EnXX  & \multirow{2}{*}{\begin{tabular}[c]{@{}l@{}}\citet{gala2023indictrans}\\ \citet{gumma-etal-2025-towards}\end{tabular}} \\
& IN22-Conv (IN22-Conv-Doc) - XXEn  &  \\
\bottomrule
\end{tabular}%
}
\caption{Evaluation datasets}
\label{tab:eval_datasets}
\end{table}

\subsection{Results}
Figure~\ref{fig:model_performance_radar} presents a comparative performance analysis of the Llama and Phi-4 architectures across NLG and NLU tasks for a diverse set of 13 Indic languages. Broadly, we observe that models fine-tuned on the \dsetname dataset (represented in green) consistently outperform existing baselines, including \bactrian, \indicalign, and \aya. This performance advantage is particularly pronounced in NLG settings, where \dsetname fine-tuned models demonstrate substantially higher performance across both high-resource languages like Hindi and Bengali, as well as lower-resource ones such as Assamese and Odia. Detailed results for the NLG tasks could be found in Table \ref{tab:nlg-tasks}. In NLU tasks, \dsetname maintains a competitive edge, often surpassing the strongest baselines highlighting the efficacy of the dataset in fostering robust multilingual understanding and generation capabilities. 

Figure \ref{fig:model_performance_scatter} illustrates the comparative performance of the models across three distinct dimensions: Understanding (NLU), Generation (NLG), and overall robustness (Win Counts). While the NLU scores (x-axis) show a competitive landscape with tight clustering among fine-tuned models, the \dsetname setting (green bubble) distinguishes itself significantly in generation tasks, consistently achieving the highest placement in terms of scores. Crucially, the bubble size indicates that \dsetname secures the highest number of `wins' - 7 for Llama3 and 8 for Phi-4 -far surpassing  other baselines like Bactrian and Aya. We observe that \dsetname has more pronounced NLU performance gains in the Phi4 setting compared to the Llama settings.

Language-wise breakdowns and dataset-level averages are reported in Appendix~\ref{detailed_results} (Tables~\ref{tab:nlu-tasks} and~\ref{tab:nlg-tasks}).

\paragraph{Evaluation on Unseen Languages}
To further understand robustness and cross-lingual transfer it is essential to test cross-lingual transfer to languages not seen in training dataset. In one of our NLG evaluation dataset(Flores) there are 16 languages which are not present in training data - Awadhi, Bhojpuri, Bodo, Chhattisgarhi, Garhwali, Haryanvi, Konkani, Maithili, Malvi, Manipuri, Marwari, Pashto, Rajasthani, Sanskrit, Santali and Tibetan. 

As illustrated in Figure~\ref{fig:updesh_comparison}, \textbf{\dsetname} consistently outperforms all the other baselines across different model architectures. On Llama-3-8B, it achieves a strong NLG score of 24.93, clearly surpassing all counterparts, and the trend holds for Phi-4, where \dsetname (24.10) again leads across baselines. These results demonstrate that \dsetname’s curation strategy enables robust cross-lingual transfer, even to languages unseen during training. Exact results on all languages could be found in Table \ref{tab:language_performance}.

\definecolor{Gray}{rgb}{0.501,0.501,0.501}

\begin{table*}[!htbp]
\small
\centering

\begin{minipage}[t]{0.48\textwidth}
\centering
\begin{tblr}{
  column{even} = {c},
  column{3} = {c},
  cell{4}{4} = {r=2}{},
  cell{6}{4} = {r=2}{},
  hline{1,8} = {-}{0.08em},
  hline{2} = {-}{0.05em},
  hline{3} = {-}{dashed,Gray},
  hline{4,6} = {-}{dashed},
}
Setting           & {NLU\\Avg} & {NLG\\Avg} & {Dataset\\Wins}                        \\
0-shot Baseline   & 28.64      & 12.69      & 0                                      \\
Updesh - R + G    & 43.97      & 32.28      & 6                                      \\
Updesh - G        & 48.48      & 19.73      & 7  \\
$\Delta$ (vs R+G) & 4.42       & -12.54     &                                        \\
Updesh - R        & 41.53      & 8.18      & 0  \\
$\Delta$ (vs R+G) & -2.43      & -24.10     &                                        
\end{tblr}

\vspace{2pt}
{\footnotesize\bfseries Llama}
\end{minipage}
\hfill
\begin{minipage}[t]{0.48\textwidth}
\centering
\begin{tblr}{
  column{even} = {c},
  column{3} = {c},
  cell{4}{4} = {r=2}{},
  cell{6}{4} = {r=2}{},
  hline{1,8} = {-}{0.08em},
  hline{2} = {-}{0.05em},
  hline{3} = {-}{dashed,Gray},
  hline{4,6} = {-}{dashed},
}
Setting           & {NLU\\Avg} & {NLG\\Avg} & {Dataset\\Wins}                        \\
0-shot Baseline   & 56.21      & 28.58      & 3                                      \\
Updesh - R + G    & 59.20      & 32.69      & 7                                      \\
Updesh - G        & 59.18      & 32.04      & 3                                      \\
$\Delta$ (vs R+G) & -0.02      & -0.64      &                                        \\
Updesh - R        & 56.79      & 1.80       & 0 \textcolor[rgb]{0.251,0.251,0.251}{} \\
$\Delta$ (vs R+G) & -2.41      & -30.86     &                                        
\end{tblr}

\vspace{2pt}
{\footnotesize\bfseries Phi-4}
\end{minipage}

\caption{Comparative performance of Updesh ablations. \textbf{R} and \textbf{G} denote the \textbf{Reasoning} and \textbf{Generation} subsets of the dataset, respectively. $\Delta$ values represent the performance difference compared to the full (R + G) model.}
\label{tab:ablations}
\end{table*}

\paragraph{Ablations on dataset composition}
To disentangle the relative contributions of the reasoning and generative subsets introduced in Sections~\ref{sec:data_gen}, we conduct controlled ablations on the \dsetname dataset. Specifically, we isolate the Reasoning (R) component - comprising translated subsets of the $\orcaagent$ and $\orcamath$ datasets focused on multi-step reasoning and chain-of-thought supervision - and the Generative (G) component, which contains open-domain instruction-following and culturally grounded synthesis tasks derived from Indic Wikipedia content.

We train individual models on each subset (\texttt{\dsetname-R} and \texttt{\dsetname-G}) and compare them against the full combined dataset (\texttt{\dsetname-R+G}) to quantify their respective effects on NLU and NLG performance. Results presented in Table~\ref{tab:ablations} reveal a clear interaction between dataset composition and model behaviour. For both Llama and Phi-4, isolating the generative subset slightly improves NLU (e.g., $+3.78$ for Llama) but substantially reduces NLG quality, suggesting that reasoning data - being of translated nature is beneficial for translation performance. Conversely, training exclusively on reasoning data leads to a marked decline in both NLU and NLG metrics ($-37.06$ for Phi-4 NLG). These results demonstrate that our bottom-up data generation approach is superior to naive translation. 

\definecolor{Gray}{rgb}{0.501,0.501,0.501}
\begin{table}
\centering
\resizebox{\linewidth}{!}{
\begin{tblr}{
  column{even} = {c},
  column{3} = {c},
  cell{4}{4} = {r=2}{},
  hline{1,6} = {-}{0.08em},
  hline{2} = {-}{0.05em},
  hline{3} = {-}{dashed,Gray},
  hline{4} = {-}{dashed},
}
Setting                        & {NLU\\Avg} & {NLG\\Avg} & {Dataset\\Wins}                        \\
0-shot Baseline                & 53.17      & 34.42      & 3                                      \\
Updesh - R + G                 & 56.61      & 38.73      & 3                                      \\
Updesh - R + G (Seq len = 32K) & 55.85      & 40.59      & 7 \textcolor[rgb]{0.251,0.251,0.251}{} \\
$\Delta$ (vs R+G)              & -0.75      & 1.86       &                                        
\end{tblr}
}
\caption{Ablations on training sequence length}
\label{tab:32k_ablations}

\end{table}
\paragraph{Ablations on training sequence length}
Additionally, given that many benchmark tasks contain shorter contexts, it is necessary to determine whether the long-context nature of the training data might have interference with short-context capabilities. Therefore, we evaluate a variant of the \dsetname trained with a 32K\footnote{halved from the original 64K} context window to examine the sensitivity of performance to sequence length constraints. Notably, we observe in \ref{tab:ablations}, that the \textbf{Phi-4 32K} variant achieves the most balanced profile, securing the highest number of dataset wins ($7$) by maintaining strong NLU performance while further boosting NLG capabilities. This indicates that training sequence length is an important design consideration.

\begin{table}[h]
    \centering
    \resizebox{\linewidth}{!}{
    \begin{tabular}{@{}l l c c c c c@{}}
        \toprule
        \textbf{Rank} & \textbf{Model} & \textbf{ELO} & \textbf{Wins} & \textbf{Losses} & \textbf{Ties} & \textbf{Win Rate} \\
        \midrule
        1 & \dsetname-32K & 1695.55 & 18760 & 6897 & 742 & 0.711 \\
        2 & \indicalign & 1658.50 & 17447 & 7991 & 657 & 0.669 \\
        3 & \dsetname & 1606.52 & 16168 & 8868 & 1111 & 0.618 \\
        4 & \dsetname Reasoning & 1537.19 & 12237 & 12371 & 1792 & 0.464 \\
        5 & \dsetname Generative & 1483.15 & 13752 & 11206 & 1442 & 0.521 \\
        6 & \bactrian & 1311.22 & 6862 & 17806 & 1554 & 0.262 \\
        7 & \textsc{aya} & 1207.86 & 2313 & 22400 & 1588 & 0.088 \\
        \bottomrule
    \end{tabular}
    }
    \caption{Overall ELO Rankings comparing \dsetname variants against baseline models for Phi-4. The \dsetname-32K model demonstrates superior performance, outperforming both internal variants and external baselines like \bactrian and \aya}
    \label{tab:elo_model_rankings}    
\end{table}

\subsection{Comparative Cultural evaluations (ELO Rankings) }
In Section~\ref{sec:model_training}, we evaluate the trained models and baselines on well-established academic benchmarks, which primarily measure performance on standardized tasks. However, such evaluations do not fully capture how useful these models are for real-world user queries spanning diverse domains, nor do they adequately reflect their helpfulness to everyday users in culturally grounded scenarios. To address this gap, it is essential to perform robust comparative evaluations in nuanced cultural contexts, focusing on non-academic, real-world questions posed by users.\footnote{Due to the high evaluation cost, we perform this analysis only on Phi-4 checkpoints; we focus on Phi-4 since it consistently outperforms Llama in our other evaluations.}

Thereby, we collect data following the collection process for Samiksha \cite{hamna2025buildingbenchmarksgroundup} and created a set of questions to assess the cultural relevance and helpfulness of LLM responses on practical, community-driven queries. Following the LLM-as-a-judge framework, we utilized \gpto to determine the superior response in pairwise comparisons between model checkpoints. Our evaluation encompasses 91,982 battles across seven models.  To ensure statistical robustness, we evaluated all possible model pairings and randomized answer positions to mitigate positional bias, and calculate ELO ratings following the work in \cite{boubdir-etal-2023-elo}. ELO ratings are calculated as the battles progress, where a higher rating indicates better comparative performance.
This evaluation indicates that \dsetname-32K significantly outperforms other baselines, with only IndicAlign showing competitive performance. Table \ref{tab:elo_model_rankings} presents a comparative evaluation of model performance based on ELO ratings. The \dsetname-32K model achieves the highest rating of 1696, establishing itself as the top-performing model in this evaluation set. It marginally outperforms Indic Align Cleaned (1659) and maintains a significant lead over the standard \dsetname (1607). Notably, the \dsetname models demonstrate substantial improvements over existing multilingual baselines, with the lead model scoring over 300 points higher than \bactrian Indic (1311) and nearly 500 points higher than the \aya Indic Sampled (1208), highlighting the efficacy of the \dsetname dataset in providing more useful and grounded answers to India-centric, domain-specific queries.

\paragraph{Alignment between ELO Scores and Automated Benchmarks} As illustrated in Figure \ref{fig:elo_analysis}, there is a clear positive correlation between the ELO scores and the evaluated metrics across the board. Notably, the \textbf{Updesh-32K (U32K)} variant demonstrates superior performance, consistently clustering in the upper-right quadrant of all three plots. It achieves the highest ELO score ($\approx$ 1700) while simultaneously maintaining leading scores in NLU Average ($\approx$ 0.55), NLG Average ($\approx$ 35), and the total number of Dataset Wins. In contrast, baseline models such as Bactrian and Aya show mixed results; while Aya remains competitive in NLG tasks, it lags significantly behind U32K in the aggregate ELO ranking.

\section{Conclusion}
In this work, we examined synthetic data as a potential remedy for the scarcity of multilingual and multicultural resources. Through a comprehensive framework and systematic experiments across Indian languages, we identified effective strategies for data generation, quality assessment, and downstream evaluation, beyond English-centric norms. We built $\dsetname$, a 9.5M 13-language IFT dataset using a culturally grounded, bottom-up pipeline. Our comprehensive evaluation spanning data generation, quality assessment (human, LLM-as-judge), and downstream tasks, revealed that synthetic data can potentially bridge resource gaps. Results show \dsetname dominates across tasks and models, for both NLU and NLG settings. We will release the $\dsetname$ dataset, evaluation protocols, and detailed analyses to enable future research.

\clearpage



\section{Limitations}
\paragraph{Lack of Reliable Data Quality Estimation for multilingual synthetic data}
Our comprehensive evaluation revealed that current LLM-based evaluators demonstrate variable reliability across quality dimensions, showing strong agreement with human judgments on objective metrics like toxicity detection (96-98\%) but significantly lower concordance on nuanced aspects like fluency assessment and persona consistency (45-60\%). This necessitates exercising caution when relying solely on LLM-based evaluations for quality estimation of multilingual synthetic data and highlights the need for more calibrated evaluators and robust evaluation frameworks.

\paragraph{Cultural Authenticity and Knowledge Base Limitations}
Cultural authenticity remains challenging due to reliance on Wikipedia as the primary knowledge base, as many cultural customs and contextual nuances specific to under-represented Indian communities may lack sufficient documentation on Wikipedia, potentially resulting in incomplete cultural representations that might favor well-documented urban practices over rural or minority contexts.

\paragraph{Lack of Specialized Benchmarking}
Limited benchmarks exist for evaluating cultural aspects and long-context/multi-turn capabilities in Indic languages, making systematic assessment of these crucial aspects difficult despite $\dsetname$'s emphasis on these capabilities. While our framework covers general NLU, NLG specialized benchmarks for cultural knowledge and reasoning are needed to systematically evaluate and make progress.
Although we conduct an comparative evaluation on real-world queries asked by human users, however we have used LLM evaluator for rating, while human evaluation would be a more fine-grained indicator of quality which we leave for future work.
\section{Ethical Considerations}

Our discussion of ethical considerations is guided by the framework proposed by \citet{bender-friedman-2018-data}.

\paragraph{Institutional Process and Oversight} The data annotation was conducted by a third-party vendor and was approved by the Institutional Review Board of our organization and by the vendor.

\paragraph{Data Provenance and Quality Assurance}
To mitigate potential artifacts and quality issues in the synthetic data, we implemented a rigorous quality control process. This process involved both automated evaluation with GPT-4o and manual verification by human annotators. We observed a high concordance between automated and human judgments on potentially problematic content. Furthermore, given that the data was generated using state-of-the-art large language models, the baseline incidence of such content was already substantially reduced. In a manual evaluation of 500 samples, human annotators flagged only 1 sample (0.2\%) on metrics pertaining to problematic content, confirming the high quality of the resulting dataset.

\paragraph{Annotator Demographics} Annotators were recruited through a professional external services company. All annotators assigned to a given data point were native speakers of the language represented in the data. Table~\ref{tab:participant-compact} summarizes the annotator demographics (education, region, age distribution, and gender).
Each data worker was compensated at the rate of \$2 per data point, which is significantly higher than the average for an regular annotation task.

\begin{table}[h]
\centering
\small
\begin{tabular}{ll}
\toprule
\textbf{Category} & \textbf{Summary} \\
\midrule
Participants & 15 \\
\midrule
Qualification & Post-graduation: 7 \\
              & Graduation: 8 \\
\midrule
Geography     & Spread across 8 Indian states \\
\midrule
Age distribution & 21--30: 7 \\
                 & 31--40: 5 \\
                 & 41--50: 3 \\
\midrule
Gender        & Female: 11 \\
              & Male: 4 \\
\bottomrule
\end{tabular}
\caption{Participant demographics summary.}
\label{tab:participant-compact}
\end{table}

\paragraph{Reproducibility} We provide a detailed reproducibility statement in Appendix~\ref{repro_statement}. 
\clearpage

\bibliography{custom,anthology-1,anthology-2}
\clearpage
\appendix
\section{Reproducibility Statement}\label{repro_statement}

To ensure full reproducibility of our pipeline—spanning data generation, model fine-tuning, downstream evaluation, and synthetic data quality assessment — we provide our complete codebase and sample generated data in the supplementary material. The submission zip contains organized data and code folders, with the code structured into seven components corresponding to distinct pipeline stages:

\begin{enumerate}
    \item \texttt{selective\_translation} – Scalable codebase for generating reasoning data through translation from $\orcaagent$.
    \item \texttt{wiki\_bt} – Wikipedia-based grounded synthetic data curation pipeline.
   \item \texttt{DCAD-2000} – Modified repository for heuristic-based filtering of standard and multi-turn conversational data.
    \item \texttt{quality\_evals} – Synthetic data quality evaluation code, including outputs, human annotations, and analysis.
    \item \texttt{model\_training} – \texttt{Axolotl}-based fine-tuning scripts for the generated datasets.
    \item \texttt{lm\_evaluation\_harness} – Custom fork integrating additional benchmarks for downstream evaluation.
    \item \texttt{plotting} – Analysis and visualization code for experimental results.
\end{enumerate}

Comprehensive instructions are provided in the main README and component-specific README files for straightforward reproduction. Model training configurations with exact hyperparameters are detailed in Appendix~\ref{tab:hyperparameters}.

We will publicly release all the following artifacts (\textbf{code and datasets}) under a permissive license.

\begin{itemize}
    \item \textbf{Complete codebase} including selective translation pipelines, Wikipedia-grounded synthetic data generation, data filtering methods, model training scripts, and the up-to-date evaluation frameworks
    \item \textbf{Training datasets} comprising \dsetname Reasoning and Generative subsets across multiple languages.
    \item \textbf{Evaluation datasets} featuring \gpto translated variants of standard benchmarks (IFEval, IFBench) for reproducibility and future benchmarking.
    \item \textbf{Raw evaluation scores} for all models across every dataset, providing complete experimental transparency
    \item \textbf{Human-annotations of synthetic data} specifically useful for calibrating / meta-evaluation of LLM evaluators on Indian languages
\end{itemize}
\begin{figure*}[htbp]
    \centering
    \includegraphics[width=1.0\linewidth]{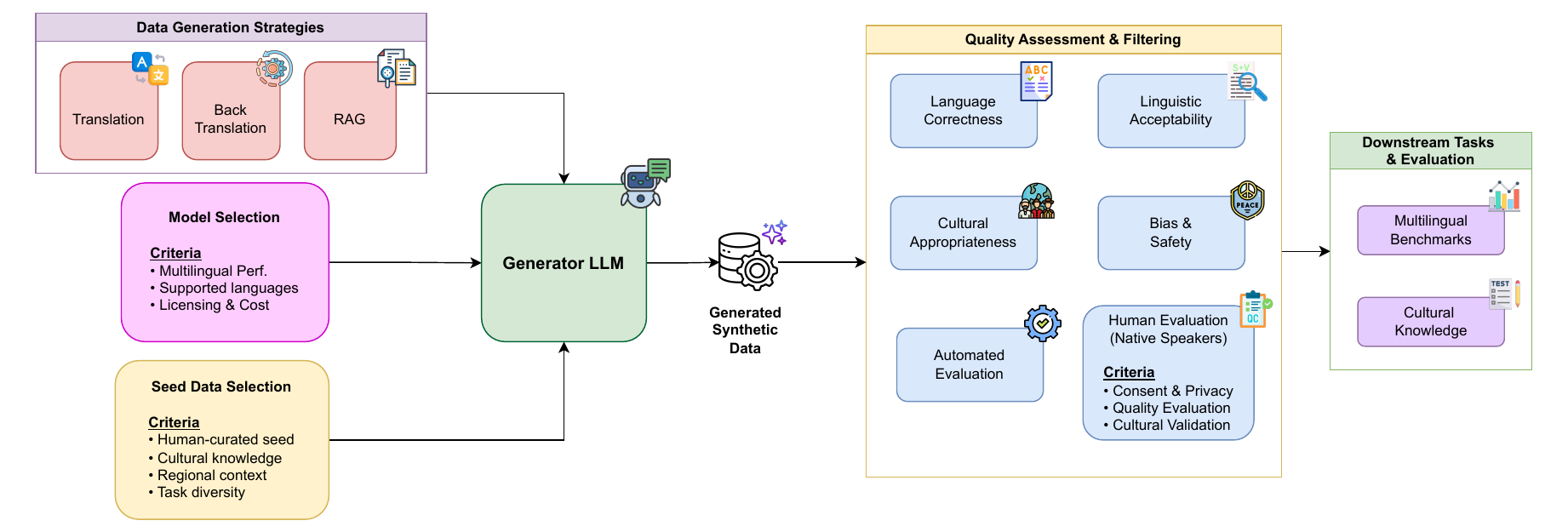}
    \caption{Framework for Multilingual \& Multicultural Synthetic Data Generation}
    \label{fig:framework}
\end{figure*}

\section{Appendix}
\label{sec:appendix}

\subsection{Design Considerations}\label{appendix:design_considerations}

Our framework (\Cref{fig:framework}) addresses synthetic data generation for multilingual and multicultural contexts throughout the AI lifecycle — pre-training, SFT, RLHF, and evaluation \citep{viswanathan-etal-2025-synthetic}. Below, we describe the key factors to consider when generating this type of data. While our focus is on SFT data, these design considerations can generalize to other synthetic data types.

\paragraph{Base model capability \& Seed data selection} Select foundation models based on performance in target languages on multilingual benchmarks. When language-specific benchmarks are unavailable, use related languages or overall multilingual performance as proxies. Other critical aspects to consider include licensing, cost, and model availability (open-weights vs. restricted). For seed data selection, cover diverse sources and tasks, prioritizing tasks containing cultural knowledge, norms, and values relevant to specific regional contexts.

\paragraph{Data generation strategy} Three primary approaches: (i) \textit{Translation} from English SFT datasets to transfer critical skills, though risking translationese artifacts \citep{zhang-toral-2019-effect,vanmassenhove-etal-2021-machine}; (ii) \textit{Back-translation}—using existing unlabeled multilingual datasets through instruction backtranslation \citep{li2024selfalignment} or back-and-forth translation \citep{nguyen-etal-2024-better}; (iii) \textit{Retrieval-augmented generation} —leveraging curated native speaker-authored content from web corpora to capture cultural knowledge and linguistic nuances. Translation-based approaches yield weaker correlations with human judgements than language-specific benchmarks \citep{kreutzer2025dj,wu2025bitterlessonlearned2000}. Bottom-up approaches grounded in web corpora show superior performance \citep{shaham-etal-2024-multilingual,khan-etal-2024-indicllmsuite,doshi-etal-2024-pretraining} but remain underexplored in multilingual contexts.

\paragraph{Quality metrics} Essential dimensions for assessment: \textit{Language correctness}—proper language, register, dialect identification \citep{marchisio-etal-2024-understanding}; \textit{Linguistic acceptability}—native speaker fluency and naturalness \citep{hada-etal-2024-large,hada-etal-2024-metal}; \textit{Cultural appropriateness}—accurate cultural references, values, and norms; \textit{Bias and safety}—absence of stereotypes and culturally inappropriate content \citep{10.1162/COLI.a.14}.

\paragraph{Downstream evaluation} Select benchmarks covering all target languages, avoiding English-translated datasets. Include diverse tasks testing cultural knowledge and values. Address benchmark contamination risks \citep{ahuja2024contaminationreportmultilingualbenchmarks} and develop new benchmarks when necessary.

\paragraph{Native speaker involvement} Engage native speakers in seed selection and evaluation. Ensure informed consent addressing cultural considerations and data sovereignty per local regulations. Exclude personally identifiable information from all sources.

\subsection{Task Descriptions}
Task Descriptions for all the subtasks in $\dsetname$ can be found in Table \ref{tab:reasoning_tasks} and \ref{tab:generative_tasks}.
\begin{table*}[!h]
\centering
\small

\begin{tabular}{lp{10cm}}
\toprule
\textbf{Task Type} & \textbf{Description} \\
\midrule
\textsc{Analytical Reasoning} & MCQ-style questions requiring step-by-step logical inference \\
\textsc{Multiple-Choice Questions} & General-purpose problems across diverse knowledge domains \\
\textsc{Fermi (Guesstimation)} & Open-ended estimation problems using logical assumptions \\
\textsc{Few-Shot Chain-of-Thought} & Tasks with 4-5 in-context examples for learning \\
\textsc{Brain Teasers} & Puzzles stimulating lateral thinking and creativity \\
\textsc{Text Classification} & Categorization tasks for predefined labels \\
\textsc{Reading Comprehension} & Questions based on understanding and interpreting textual passages \\
\textsc{Math} & Grade-school arithmetic, algebra, and geometry word problems \\
\bottomrule
\end{tabular}

\caption{Reasoning task categories and their descriptions.}
\label{tab:reasoning_tasks}
\end{table*}
\begin{table*}[htbp]
\centering
\small
\begin{tabular}{lp{5.5cm}p{3.5cm}c}
\toprule
\textbf{Task Type} & \textbf{Synthesis Method} & \textbf{Phases} & \textbf{Qwen3-Mode} \\
\midrule
\textsc{Logical Reasoning} & Generate implicit inferences from text passages & (1) Direct inference generation & Reasoning \\
\midrule
\textsc{Multi-Hop QA} & Create questions requiring information synthesis across text segments & (1) Question generation (2) Answer generation & Reasoning \\
\midrule
\textsc{Creative Writing} & Transform factual content into engaging narratives & (1) Generate creative piece (2) Generate eliciting prompt & Reasoning \\
\midrule
\textsc{Multi-Turn Dialogue} & Agentic workflows with 3-5 turn conversations between personas & (1) Generate dialog adhering to personas (2) Generate natural prompt & Non-reasoning \\
\midrule
\textsc{Summarization} & Generate summaries preserving key information across languages & (1) Direct summary generation & Non-reasoning \\
\midrule
\textsc{Machine Translation} & Cross-lingual conversion maintaining cultural context & (1) Direct translation & Non-reasoning \\
\midrule
\textsc{Causal Reasoning} & Identify and explain cause-effect relationships in text & (1) Direct causal analysis & Reasoning \\
\bottomrule
\end{tabular}

\caption{Generative task categories with synthesis methods, phases, and model configuration}
\label{tab:generative_tasks}
\end{table*}

\clearpage

\subsection{Rubrics used for the quality evaluation of the synthetic data}\label{tab:rubrics}

\noindent\textbf{Creative Writing}
\begin{itemize}[leftmargin=*]
    \item \textit{Instruction adherence}: Assesses if the output strictly follows all constraints and guidelines provided in the prompt.
    \item \textit{Fluency}: Evaluates the naturalness, grammatical correctness, and readability of the generated text.
    \item \textit{Narrative coherence}: Checks for logical consistency in the plot, character development, and thematic elements.
\end{itemize}

\noindent\textbf{Reasoning Tasks}
\begin{itemize}[leftmargin=*]
    \item \textit{Answer adequacy}: Determines if the final answer is correct, complete, and directly addresses the core question.
    \item \textit{Context adherence}: Measures whether the reasoning remains faithful to the provided context, avoiding external facts.
    \item \textit{Instruction adherence}: Verifies that the output's structure, format, and steps match the user's instructions.
    \item \textit{Fluency and readability}: Assesses the clarity, logical flow, and ease of understanding of the explanation.
    \item \textit{Problematic content and cultural relevance}: Scrutinizes the response for harmful stereotypes and ensures it is culturally sensitive and appropriate.
\end{itemize}

\noindent\textbf{Multi-turn Dialog}
\begin{itemize}[leftmargin=*]
    \item \textit{Persona adherence}: Evaluates the model's ability to consistently maintain a specific character or role throughout the conversation.
    \item \textit{Topic adherence}: Checks if the conversation remains focused on the established topic or transitions logically.
    \item \textit{Linguistic plausibility}: Assesses whether the dialogue sounds natural, human-like, and contextually appropriate.
    \item \textit{Repetitiveness}: Measures the degree to which the model avoids unnecessarily repeating phrases or ideas.
    \item \textit{Toxicity check}: Ensures the response is free from any offensive, harmful, or inappropriate content.
    \item \textit{Instruction adherence}: Verifies that the model follows meta-instructions given by the user during the dialogue.
\end{itemize}

\noindent\textbf{Summarization}
\begin{itemize}[leftmargin=*]
    \item \textit{Coverage}: Determines if the summary successfully captures all essential points from the source text.
    \item \textit{Factual accuracy}: Checks that the summary correctly represents the information and facts from the original document.
    \item \textit{Conciseness}: Evaluates whether the summary is significantly shorter than the source while retaining critical information.
    \item \textit{Coherence and logical flow}: Assesses if the summary is well-structured, logically organized, and easy to follow.
    \item \textit{Style and tone}: Measures how well the summary reflects the style and tone of the original text.
\end{itemize}

\noindent\textbf{Translation}
\begin{itemize}[leftmargin=*]
    \item \textit{Semantic correctness}: Assesses whether the meaning, intent, and nuance of the source text are accurately conveyed.
    \item \textit{Fluency correctness}: Evaluates the grammatical accuracy and naturalness of the translated text in the target language.
    \item \textit{Domain appropriateness}: Checks if the terminology is correct and suitable for the specific subject matter (e.g., legal, medical).
    \item \textit{Style and tone}: Determines if the translation successfully captures the original author's writing style and emotional tone.
    \item \textit{Completeness}: Verifies that the entire source text has been translated without any omissions or additions.
\end{itemize}

\clearpage
\subsection{Results from the quality evaluations of the synthetic data} \label{sec:quality_eval}
\begin{figure*}[!htbp]
    \centering
    \includegraphics[width=1\linewidth]{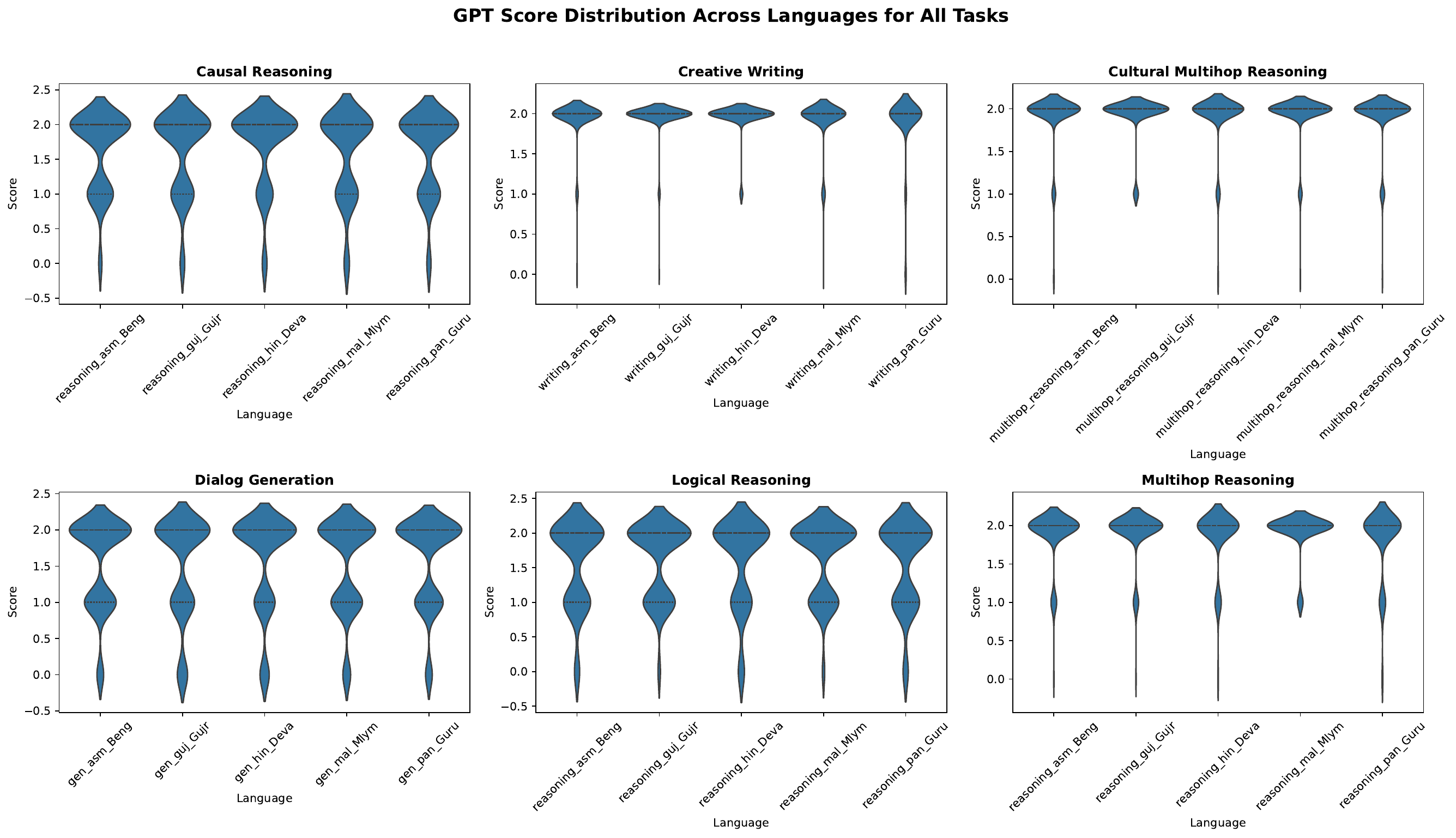}
    \caption{LLM evaluations across 5 synthetically generated tasks}
    \label{fig:llm_evals}
\end{figure*}

\begin{figure*}[!htbp]
    \centering
    \includegraphics[width=1\linewidth]{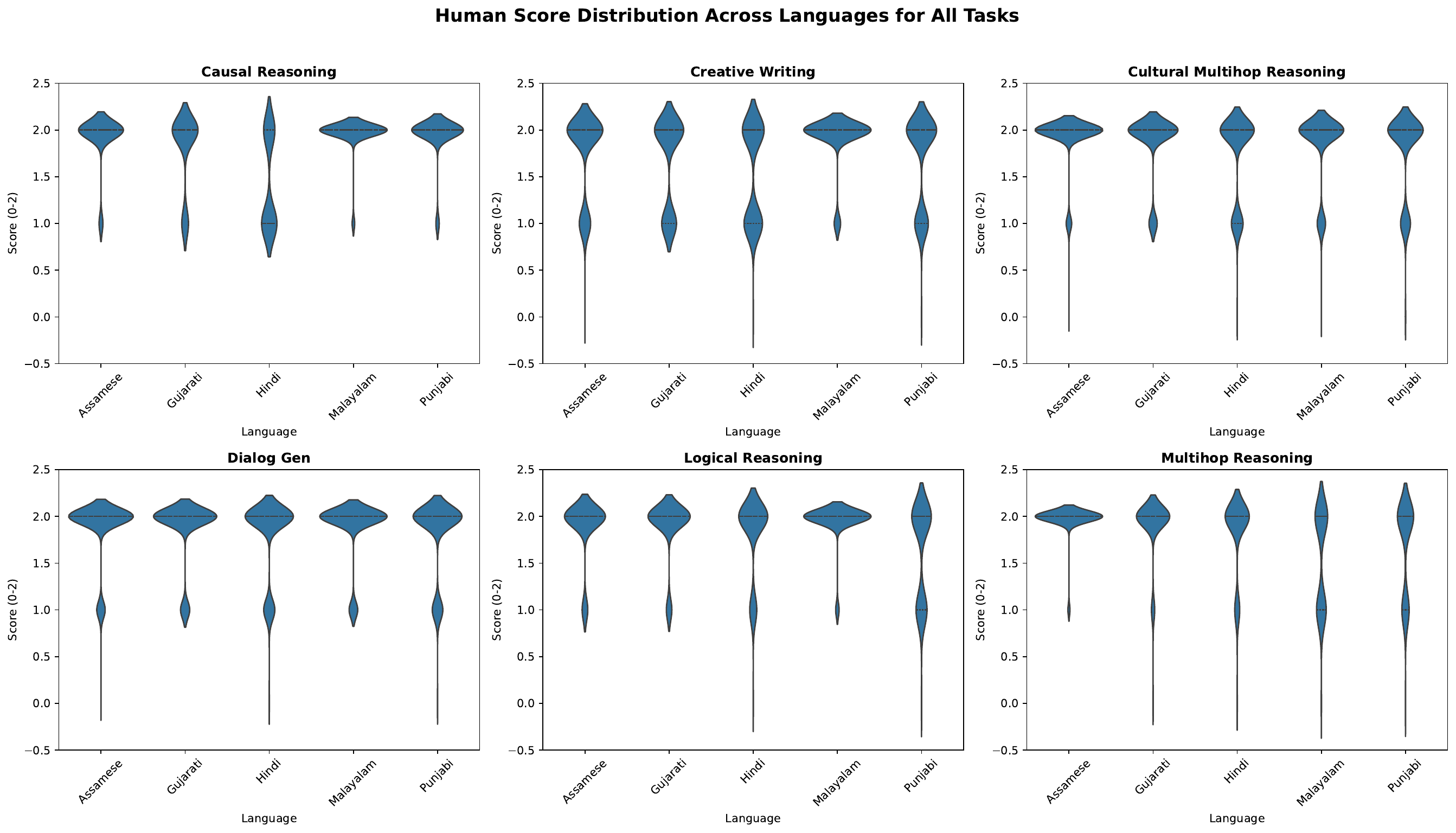}
    \caption{Expert human evaluations across 5 synthetically generated tasks}
    \label{fig:human_evals}
\end{figure*}

Figures ~\ref{fig:llm_evals} and \ref{fig:human_evals} show the distribution of the scores received for the tasks we evaluated across the languages for the tasks. Expert human evaluators have consistently given a score of 2 across languages and tasks, indicating the high quality of \dsetname. The disagreements between humans and the LLM is however unclear from these plots. We thereby, performed a thorough inter annotator analysis, the details of which are present in the next figures.

\begin{figure*}[!htbp]
    \centering
    \includegraphics[width=\linewidth]{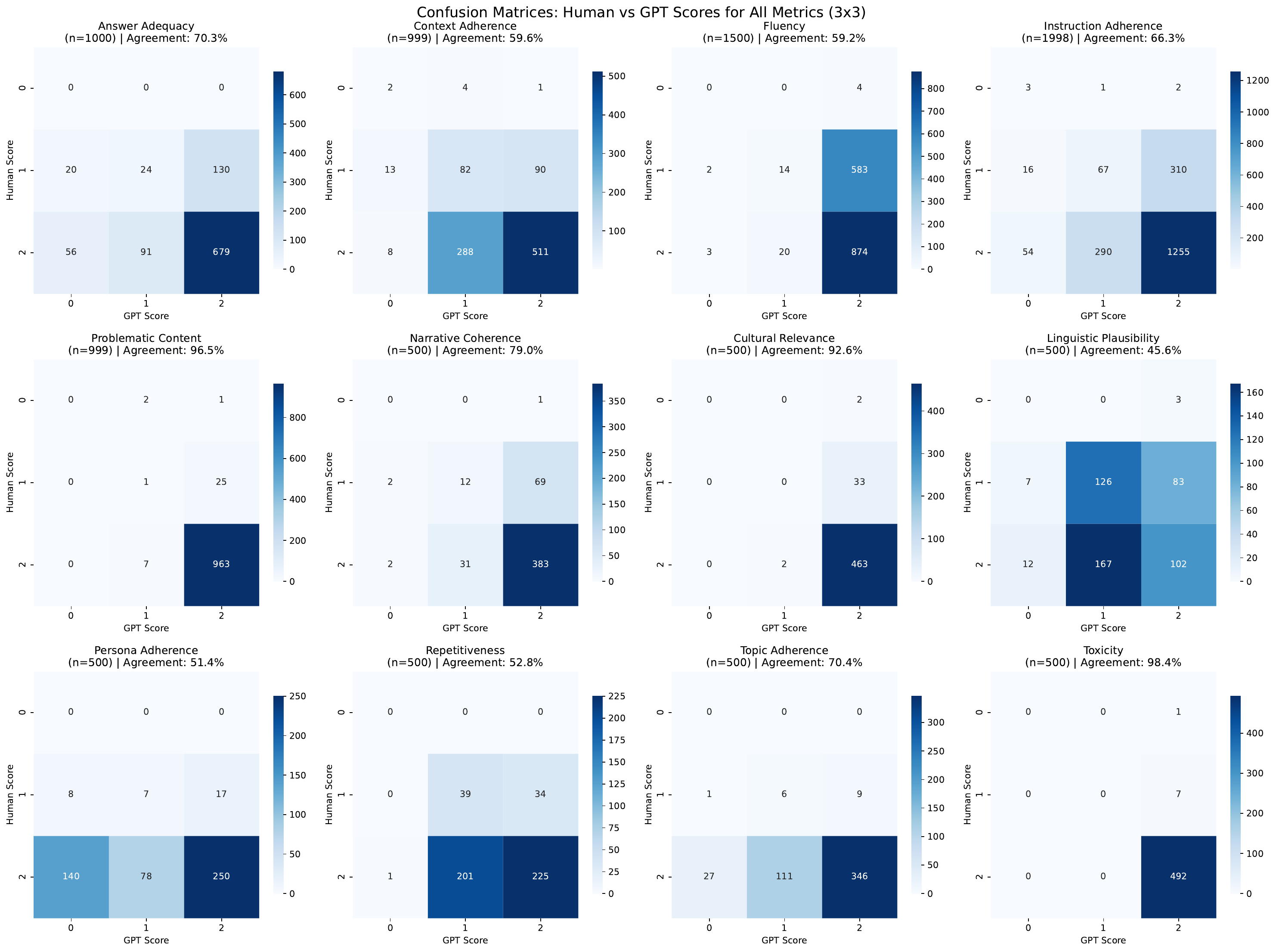}
    \caption{Confusion matrices showing agreement between human and LLM evaluators}
    \label{fig:confusion_matrices}
\end{figure*}

\begin{figure*}[!htbp]
    \centering
    \includegraphics[width=\linewidth]{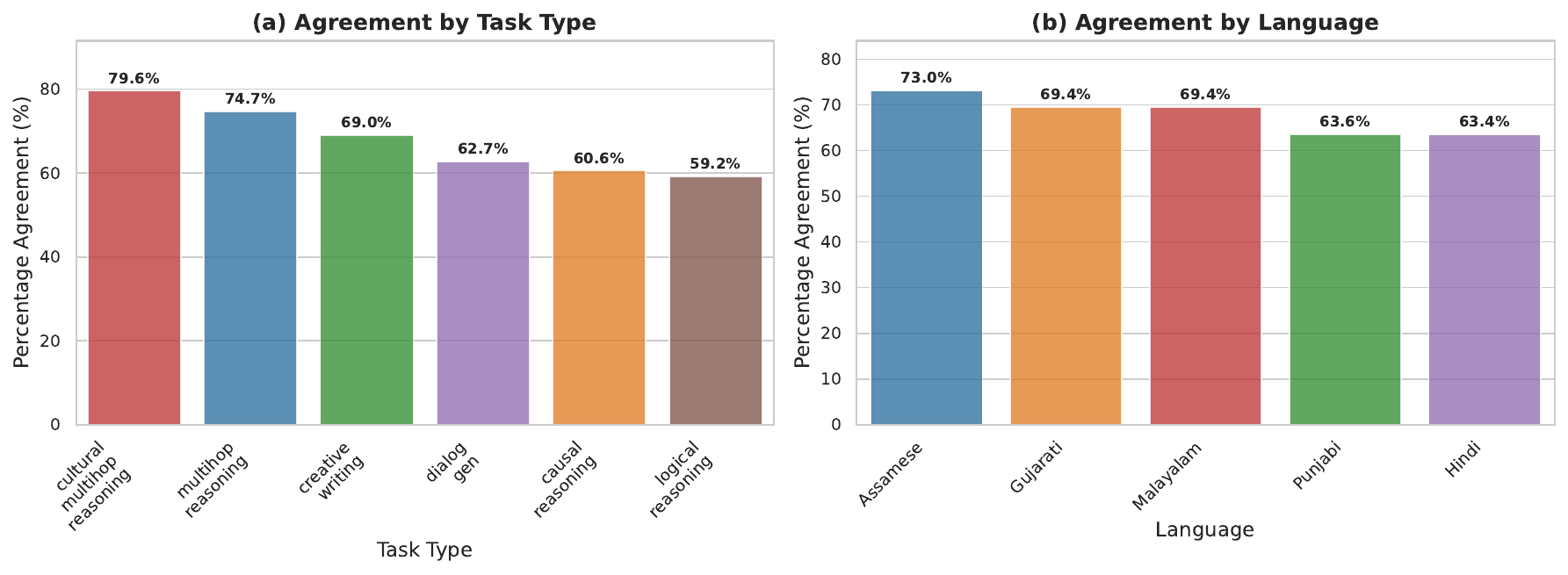}
    \caption{Agreement between human and LLM evaluators per task and language, respectively}
    \label{fig:agreement_analysis}
\end{figure*}

Continuing the claims made from Figure \ref{fig:metric_agreement}, Figure \ref{fig:confusion_matrices} provides a clearer view of the tasks on which humans and LLMs are likely to agree or disagree. We find that the largest disagreements occur in tasks such as assessing the linguistic plausibility of a given text in a regional language. Furthermore, LLMs struggle with evaluating long-context tasks, such as evaluating whether the same persona is maintained throughout a multi-turn conversation in a regional language. There is also a notable divergence between what human evaluators consider fluent in a relatively low-resource language and what an LLM deems fluent. In contrast, we observe considerable agreement in tasks like toxicity detection and problematic content flagging. LLMs also perform reasonably well at identifying whether a text is culturally relevant, but issues arise when the evaluation requires more fine-grained judgments of multilinguality and multiculturalism. We do not identify any specific trends language or task-wise as apparent in \Cref{fig:agreement_analysis}.

\definecolor{thinking_color}{RGB}{194, 213, 247}


\subsection{Generation Hyperparameters}
For data synthesis, decoding is performed using nucleus sampling with $top\_p = 0.95$ and $temperature=1.0$.

\subsection{Training Hyperparameters}
Hyperparameters for all our training runs could be found in Table \ref{tab:hyperparameters}
\label{app:hyperparameters}
\begin{table}[!htbp]
\centering
\small
\resizebox{\columnwidth}{!}{%
\begin{tabular}{lc}
\toprule
\textbf{Hyperparameter} & \textbf{Value} \\
\midrule
Base Model & \texttt{phi4-base} / \texttt{llama-3.1-8b} \\
Sequence Length & 65,536 \\
Effective Batch Size & 8192 \\
Number of Epochs & 3 \\
Optimizer & AdamW \\
Learning Rate & 1e-5 \\
LR Scheduler & Cosine \\
Adam Betas & (0.9, 0.95) \\
Max Grad Norm & 1.0 \\
Warmup Ratio & 0.03 \\
Weight Decay & 0.1 \\
NEFTune Noise Alpha \cite{jain2024neftune} & 5 \\
Precision & BF16 \\
Flash Attention & True \\
Gradient Checkpointing & True \\
\bottomrule
\end{tabular}%
}
\caption{Training hyperparameters used for all our experiments.}
\label{tab:hyperparameters}
\end{table}

\subsection{Data Quality Assesment of Reasoning Data}

\begin{table}[H]
\centering
\resizebox{\columnwidth}{!}{%
\begin{tabular}{lcccccccc}
\toprule
\textbf{Language} & \textsc{\textbf{Analytical}} & \textsc{\textbf{Brain}} & \textsc{\textbf{Fermi}} & \textsc{\textbf{Fs-CoT}} & \textsc{\textbf{Math}} & \textsc{\textbf{MCQ}} & \textsc{\textbf{RC}} & \textsc{\textbf{Text Class}} \\
\midrule
Assamese & 75.02 & 71.62 & 79.93 & 96.59 & 79.87 & 64.33 & 65.07 & 71.29 \\
Bengali & 87.25 & 77.40 & 80.69 & 82.10 & 79.87 & 67.25 & 74.94 & 74.72 \\
Gujarati & 77.86 & 67.15 & 82.14 & 73.96 & 49.66 & 78.04 & 63.21 & 55.95 \\
Hindi & 84.49 & 79.23 & 81.54 & 87.11 & 64.25 & 74.80 & 73.71 & 67.00 \\
Kannada & 79.96 & 76.87 & 80.02 & 81.26 & 65.80 & 69.91 & 64.15 & 60.77 \\
Malayalam & 75.21 & 73.41 & 70.10 & 77.93 & 68.51 & 55.69 & 63.19 & 75.63 \\
Marathi & 77.84 & 68.63 & 69.33 & 82.81 & 68.56 & 64.48 & 56.49 & 60.58 \\
Nepali & 81.79 & 86.18 & 74.71 & 83.32 & 53.98 & 56.42 & 53.86 & 59.63 \\
Odia & 56.47 & 62.06 & 50.70 & 93.61 & 57.55 & 51.21 & 42.95 & 52.15 \\
Punjabi & 83.75 & 51.79 & 77.40 & 79.04 & 61.19 & 51.21 & 69.81 & 54.11 \\
Tamil & 79.28 & 70.70 & 74.54 & 75.16 & 57.83 & 60.66 & 63.94 & 49.65 \\
Telugu & 78.24 & 80.26 & 74.33 & 79.91 & 69.88 & 60.66 & 61.93 & 60.95 \\
Urdu & 85.05 & 79.97 & 66.31 & 81.93 & 59.80 & 61.76 & 64.91 & 67.23 \\
\bottomrule
\end{tabular}%
}
\caption{Backtranslation ChrF scores for the Reasoning subset.}
\label{tab:qc_backtranslation_reasoning}
\end{table}


\clearpage
\onecolumn
\newtcolorbox{PromptBox}[2][]{
  title=\textbf{#2},
  breakable,
  fonttitle=\bfseries,
  enhanced,
  colback=thinking_color!10,   
  colbacktitle=thinking_color, 
  coltitle=black,              
  colframe=black,              
  coltext=black,               
  boxrule=0.5pt,
  arc=2mm,
  #1
}

\subsection{Detailed Results}\label{detailed_results} 

\begin{table}[h]
\centering
\small
\resizebox{\textwidth}{!}{%
\begin{tabular}{llcccccccccc} 
\toprule
\textbf{Model} & \textbf{Setting} & \textbf{NLU Avg} & \textbf{MMLU-I} & \textbf{MILU} & \textbf{ARC-I} & \textbf{BoolQ-I} & \textbf{TVQA-I} & \textbf{Bele} & \textbf{INCL} & \textbf{GMMLU} \\
\midrule
\multirow{5}{*}{\textbf{Llama 8B}} 
& \textbf{Base}        & 29.33 & 22.97 & 25.88 & 25.26 & 62.04 & 26.55 & 24.06 & 24.88 & 23.02 \\
& \textbf{\bactrian}    & \textbf{45.37} & \textbf{36.80} & 39.41 & 26.49 & 71.21 & \textbf{66.30} & 41.45 & \textbf{39.14} & \textbf{42.14} \\
& \textbf{\indicalign}  & 42.14 & 34.47 & \textbf{39.67} & 27.35 & 71.14 & 59.35 & 32.09 & 34.37 & 38.67 \\
& \textbf{\aya}         & 42.73 & 32.69 & 36.98 & 28.11 & 76.24 & 50.65 & 46.54 & 33.84 & 36.81 \\
& \textbf{\dsetname}      & 42.89 & 34.40 & 36.13 & \textbf{29.67} & \textbf{77.51} & 46.13 & \textbf{51.30} & 30.56 & 37.43 \\
\midrule
\multirow{5}{*}{\textbf{Phi-4}} 
& \textbf{Base}        & 53.17 & 43.38 & 49.22 & 29.99 & 79.56 & 68.25 & 54.55 & 45.46 & 54.97 \\
& \textbf{\bactrian}    & 49.25 & 39.91 & 45.09 & 29.40 & 74.40 & 67.30 & 48.32 & 41.72 & 47.89 \\
& \textbf{\indicalign}  & 51.71 & 44.99 & 48.25 & 32.23 & 76.26 & 61.68 & 51.68 & 44.71 & 53.92 \\
& \textbf{\aya}         & 54.41 & 45.08 & 50.57 & \textbf{33.01} & 72.62 & \textbf{71.80} & 60.71 & \textbf{48.02} & 53.47 \\
& \textbf{\dsetname}      & \textbf{56.61} & \textbf{48.98} & \textbf{51.19} & 32.08 & \textbf{81.38} & 60.14 & \textbf{74.42} & 47.24 & \textbf{57.47} \\
\bottomrule
\end{tabular}%
}
\caption{Performance comparison of Llama 8B and Phi-4 variants across Indic NLU Tasks. All entries are \emph{accuracy} (higher is better)}
\label{tab:nlu-tasks}
\end{table}

\begin{table}[h!]
\centering
\small
\begin{tabular}{llcccccc} 
\toprule
\textbf{Model} & \textbf{Setting} & \textbf{\makecell{NLG\\Avg}} & \textbf{\makecell{Flores\\En-XX}} & \textbf{\makecell{Flores\\XX-En}} & \textbf{XSum} & \textbf{\makecell{IN22-Conv-Doc\\En-XX}} & \textbf{\makecell{IN22-Conv-Doc\\XX-En}} \\
\midrule
\multirow{5}{*}{\textbf{Llama 8B}} 
& \textbf{Base} & 8.91  & 1.45  & 41.71 & 0.16  & 0.60  & 0.60 \\
& \textbf{\bactrian} & 23.81 & 28.85 & 50.98 & 0.21  & 19.51 & 19.51 \\
& \textbf{\indicalign} & 15.67 & 32.12 & 3.20  & 12.84 & 15.09 & 15.09 \\
& \textbf{\aya} & 9.69  & 28.38 & 0.46  & 0.23  & 9.69  & 9.69 \\
& \textbf{\dsetname} & \textbf{36.41} & \textbf{44.00} & \textbf{51.88} & \textbf{25.54} & \textbf{30.31} & \textbf{30.31} \\
\midrule
\multirow{5}{*}{\textbf{Phi4}} 
& \textbf{Base} & 34.20 & 30.23 & 56.57 & 17.59 & \textbf{33.31} & \textbf{33.31} \\
& \textbf{\bactrian} & 24.45 & 26.78 & 51.86 & 0.31  & 21.60 & 21.71 \\
& \textbf{\indicalign} & 13.77 & 32.13 & 0.59  & 0.28  & 17.89 & 17.94 \\
& \textbf{\aya} & 13.93 & 30.16 & 1.58  & 0.37  & 18.77 & 18.78 \\
& \textbf{\dsetname} & \textbf{36.56} & \textbf{45.82} & \textbf{59.55} & \textbf{21.66} & 27.81 & 27.94 \\
\bottomrule
\end{tabular}
\caption{Performance comparison of Llama 8B and Phi-4 variants across Indic NLG. All entries are \emph{ChrF} scores (higher is better).}
\label{tab:nlg-tasks}
\end{table}

\begin{table}[h!]
\centering
\setlength{\tabcolsep}{2.5pt}
\resizebox{\textwidth}{!}{%
\begin{tabular}{llccccccccccccccc}
\toprule
\textbf{Model} & \textbf{Variant} & \textbf{Avg} & \textbf{as} & \textbf{bn} & \textbf{en} & \textbf{gu} & \textbf{hi} & \textbf{kn} & \textbf{ml} & \textbf{mr} & \textbf{ne} & \textbf{or} & \textbf{pa} & \textbf{ta} & \textbf{te} & \textbf{ur} \\
\midrule

\multirow{5}{*}{\textbf{Llama 8B}} 
 & Base ZS    & 28.98 & 22.22 & 29.61 & 36.08 & 29.97 & 29.19 & 30.41 & 30.82 & 30.72 & 23.02 & 31.14 & 30.14 & 29.91 & 28.97 & 23.44 \\
 & Bactrian   & 45.65 & 38.22 & 45.73 & 68.19 & 44.06 & 50.88 & 43.41 & 44.07 & 48.05 & 40.12 & 40.24 & 44.90 & 40.81 & 40.35 & 50.11 \\
 & IndicAlign & 41.03 & 32.11 & 42.04 & 63.79 & 42.05 & 44.55 & 42.36 & 41.30 & 43.38 & 31.84 & 38.87 & 41.52 & 40.79 & 39.83 & 30.00 \\
 & Aya        & 44.30 & 42.78 & 42.43 & 63.04 & 42.07 & 45.66 & 43.28 & 43.76 & 43.15 & 36.03 & 38.52 & 43.06 & 41.74 & 40.85 & 53.89 \\
 & Updesh     & 45.23 & 46.44 & 41.97 & 59.84 & 44.52 & 42.35 & 44.95 & 45.56 & 42.99 & 35.63 & 43.99 & 44.08 & 41.82 & 41.98 & 57.11 \\
\midrule \addlinespace

\multirow{5}{*}{\textbf{Phi-4}}    
 & Base ZS    & 53.36 & 42.22 & 52.77 & 75.64 & 53.93 & 60.37 & 51.55 & 49.40 & 53.25 & 48.14 & 47.51 & 54.21 & 48.00 & 47.84 & 62.22 \\
 & Bactrian   & 50.16 & 40.67 & 49.39 & 77.47 & 49.21 & 56.36 & 45.57 & 45.53 & 50.54 & 45.06 & 44.47 & 50.60 & 41.87 & 41.80 & 63.67 \\
 & IndicAlign & 52.86 & 49.33 & 51.13 & 75.84 & 50.74 & 55.49 & 50.33 & 50.88 & 49.38 & 49.91 & 42.72 & 51.22 & 48.07 & 47.44 & 67.56 \\
 & Aya        & 55.80 & 52.89 & 53.41 & 76.41 & 55.16 & 60.23 & 54.50 & 51.63 & 55.01 & 52.87 & 46.25 & 55.46 & 50.10 & 49.33 & 68.00 \\
 & Updesh     & 59.71 & 70.33 & 55.77 & 75.45 & 56.91 & 61.82 & 56.43 & 54.20 & 56.81 & 58.21 & 53.35 & 57.59 & 51.49 & 52.04 & 75.56 \\

\bottomrule
\end{tabular}%
}
\caption{\textbf{NLU Performance}: Evaluation across Indic languages including English. Language codes: \textbf{as}: Assamese, \textbf{bn}: Bengali, \textbf{en}: English, \textbf{gu}: Gujarati, \textbf{hi}: Hindi, \textbf{kn}: Kannada, \textbf{ml}: Malayalam, \textbf{mr}: Marathi, \textbf{ne}: Nepali, \textbf{or}: Odia, \textbf{pa}: Punjabi, \textbf{ta}: Tamil, \textbf{te}: Telugu, \textbf{ur}: Urdu.}
\label{tab:nlu_perf_langwise}
\end{table}

\begin{table}[h!]
\centering
\setlength{\tabcolsep}{2.5pt}
\resizebox{\textwidth}{!}{%
\begin{tabular}{llcccccccccccccc}
\toprule
\textbf{Model} & \textbf{Variant} & \textbf{Avg} & \textbf{as} & \textbf{bn} & \textbf{gu} & \textbf{hi} & \textbf{kn} & \textbf{ml} & \textbf{mr} & \textbf{ne} & \textbf{or} & \textbf{pa} & \textbf{ta} & \textbf{te} & \textbf{ur} \\
\midrule

\multirow{5}{*}{\textbf{Llama 8B}} 
 & Base ZS    & 9.08  & 7.05 & 7.87 & 7.76 & 11.05 & 9.22 & 9.04 & 9.77 & 11.83 & 8.48 & 10.26 & 7.30 & 9.24 & 9.22 \\
 & Bactrian   & 24.24 & 20.32 & 29.79 & 24.24 & 33.13 & 22.87 & 23.97 & 19.58 & 28.30 & 18.26 & 21.63 & 24.24 & 23.32 & 25.51 \\
 & IndicAlign & 15.80 & 13.09 & 18.29 & 12.27 & 26.29 & 15.34 & 14.75 & 17.43 & 21.57 & 6.65 & 13.71 & 17.22 & 14.66 & 14.14 \\
 & Aya        & 9.87  & 7.49 & 12.98 & 10.83 & 8.62 & 11.07 & 12.74 & 8.04 & 11.86 & 6.42 & 8.34 & 12.00 & 8.78 & 9.10 \\
 & Updesh     & 36.68 & 33.80 & 35.98 & 34.90 & 36.81 & 35.73 & 37.75 & 34.15 & 43.13 & 32.87 & 35.60 & 39.08 & 39.64 & 37.42 \\
\midrule \addlinespace

\multirow{5}{*}{\textbf{Phi-4}}    
 & Base ZS    & 34.56 & 25.39 & 39.47 & 38.83 & 46.31 & 30.02 & 30.97 & 32.36 & 40.76 & 27.36 & 40.32 & 30.73 & 34.61 & 32.10 \\
 & Bactrian   & 24.96 & 20.74 & 30.75 & 24.90 & 40.36 & 22.72 & 23.68 & 23.53 & 33.54 & 13.29 & 17.42 & 25.19 & 25.27 & 23.14 \\
 & IndicAlign & 14.08 & 12.81 & 16.57 & 15.35 & 25.27 & 12.46 & 11.59 & 13.06 & 20.42 & 6.95 & 10.81 & 15.58 & 19.52 & 2.60 \\
 & Aya        & 14.29 & 7.67 & 21.88 & 15.44 & 25.36 & 5.15 & 12.57 & 12.80 & 23.35 & 1.91 & 16.39 & 12.33 & 15.49 & 15.40 \\
 & Updesh     & 36.87 & 32.60 & 37.66 & 38.50 & 40.38 & 34.21 & 36.50 & 35.12 & 42.16 & 33.56 & 36.29 & 39.99 & 37.61 & 34.74 \\

\bottomrule
\end{tabular}%
}
\caption{\textbf{NLG Performance}: Evaluation across Indic languages (excluding English). Language codes: \textbf{as}: Assamese, \textbf{bn}: Bengali, \textbf{gu}: Gujarati, \textbf{hi}: Hindi, \textbf{kn}: Kannada, \textbf{ml}: Malayalam, \textbf{mr}: Marathi, \textbf{ne}: Nepali, \textbf{or}: Odia, \textbf{pa}: Punjabi, \textbf{ta}: Tamil, \textbf{te}: Telugu, \textbf{ur}: Urdu.}
\label{tab:nlg_perf_langwise}
\end{table}


\begin{table}[h!]
\centering
\setlength{\tabcolsep}{2.5pt} 
\resizebox{\textwidth}{!}{%
\begin{tabular}{llccccccccccccccccc}
\toprule
\textbf{Model} & \textbf{Variant} & \textbf{Avg} & \textbf{awa} & \textbf{bho} & \textbf{brx} & \textbf{hne} & \textbf{gbm} & \textbf{bgc} & \textbf{gom} & \textbf{mai} & \textbf{mup} & \textbf{mni} & \textbf{mwr} & \textbf{ps} & \textbf{hoj} & \textbf{sa} & \textbf{sat} & \textbf{bo} \\
\midrule

\multirow{5}{*}{\textbf{Llama 8B}} 
 & Base ZS    & 10.13 & 15.86 & 14.82 & 2.64 & 13.06 & 13.72 & 15.41 & 7.24 & 9.34 & 13.66 & 1.00 & 15.13 & 13.29 & 13.88 & 5.81 & 2.80 & 4.39 \\
 & Bactrian   & 19.59 & 27.70 & 19.28 & 4.45 & 28.41 & 27.56 & 26.20 & 13.23 & 23.72 & 26.71 & 5.00 & 29.27 & 20.09 & 26.89 & 17.38 & 4.25 & 13.29 \\
 & IndicAlign & 9.19 & 12.45 & 11.03 & 0.32 & 15.52 & 11.10 & 15.35 & 6.00 & 10.16 & 13.79 & 2.23 & 14.36 & 7.14 & 14.43 & 11.59 & 0.21 & 1.38 \\
 & Aya        & 6.10 & 6.97 & 7.63 & 0.24 & 7.28 & 8.87 & 9.23 & 4.28 & 7.16 & 8.93 & 2.16 & 9.36 & 3.16 & 9.78 & 6.55 & 0.23 & 5.82 \\
 & Updesh     & 24.40 & 33.67 & 32.33 & 4.30 & 37.34 & 30.20 & 35.61 & 23.53 & 26.23 & 33.13 & 6.60 & 36.50 & 21.46 & 33.69 & 21.25 & 4.55 & 10.04 \\
\midrule \addlinespace

\multirow{5}{*}{\textbf{Phi-4}}    
 & Base ZS    & 22.69 & 35.45 & 32.75 & 5.01 & 30.05 & 31.04 & 26.78 & 15.74 & 31.99 & 27.04 & 5.73 & 29.62 & 25.09 & 24.09 & 26.08 & 4.99 & 11.61 \\
 & Bactrian   & 22.19 & 31.71 & 27.96 & 4.80 & 31.14 & 30.38 & 31.70 & 15.38 & 26.43 & 30.58 & 5.90 & 31.13 & 19.33 & 30.11 & 21.13 & 3.80 & 13.64 \\
 & IndicAlign & 9.41 & 12.51 & 10.26 & 1.33 & 12.83 & 12.65 & 12.83 & 9.09 & 17.01 & 13.51 & 1.28 & 13.52 & 6.76 & 10.14 & 11.03 & 2.21 & 3.58 \\
 & Aya        & 12.20 & 17.67 & 13.22 & 1.12 & 16.44 & 16.89 & 17.25 & 7.98 & 18.41 & 14.90 & 0.55 & 16.13 & 12.47 & 14.14 & 14.86 & 4.25 & 8.95 \\
 & Updesh     & 23.51 & 34.32 & 30.71 & 4.90 & 35.04 & 32.39 & 28.16 & 21.88 & 26.82 & 31.22 & 7.29 & 33.82 & 21.04 & 29.12 & 22.06 & 4.19 & 13.15 \\

\bottomrule
\end{tabular}%
}
\caption{Detailed performance breakdown for out-of-training languages. Language codes: \textbf{awa}: Awadhi, \textbf{bho}: Bhojpuri, \textbf{brx}: Bodo, \textbf{hne}: Chhattisgarhi, \textbf{gbm}: Garhwali, \textbf{bgc}: Haryanvi, \textbf{gom}: Konkani, \textbf{mai}: Maithili, \textbf{mup}: Malvi, \textbf{mni}: Manipuri, \textbf{mwr}: Marwari, \textbf{ps}: Pashto, \textbf{hoj}: Rajasthani, \textbf{sa}: Sanskrit, \textbf{sat}: Santali, \textbf{bo}: Tibetan.}
\label{tab:language_performance}
\end{table}

\begin{figure*}[htbp]
    \centering
    \includegraphics[width=\textwidth]{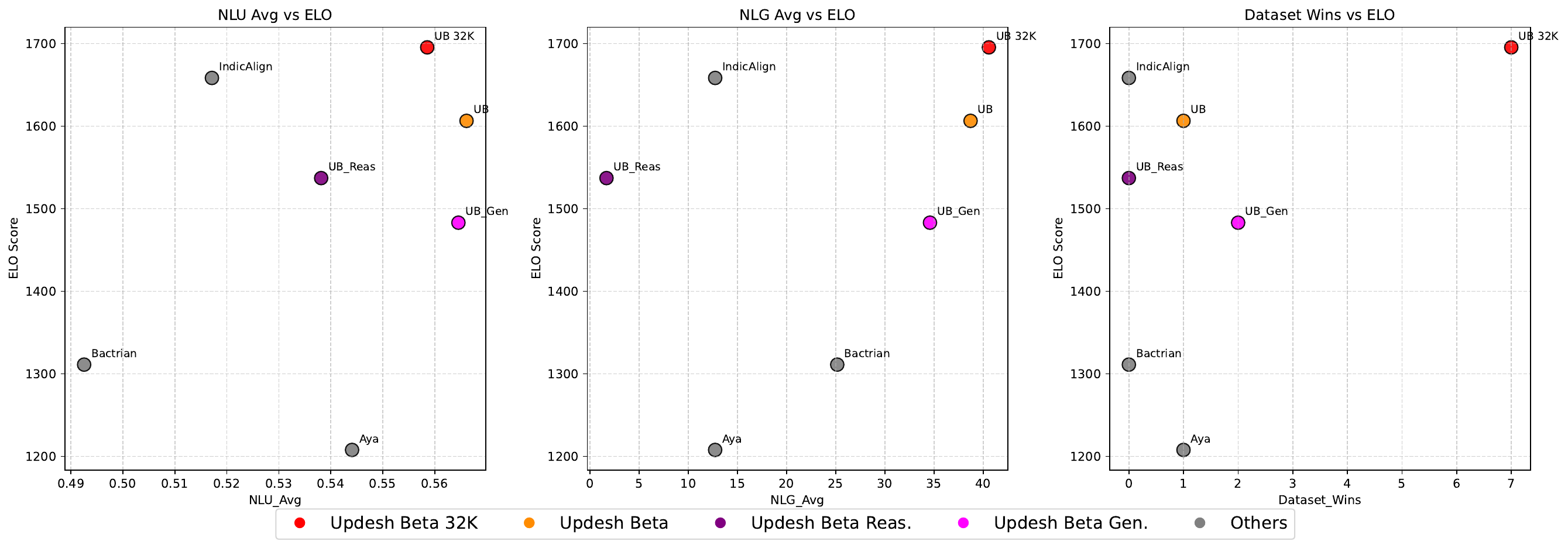} 
    \caption{Scatter plots correlating ELO Scores with NLU Average (left), NLG Average (center), and Dataset Wins (right). The Updesh-32K model (UB 32K) consistently outperforms baselines, appearing in the top-right quadrant across all metrics.}
    \label{fig:elo_analysis}
\end{figure*}

\clearpage
\section{Cultural Evaluation Framework}
\label{Cult-evals}
\begin{figure*}[!htbp]
    \centering
    \includegraphics[width=0.90\linewidth]{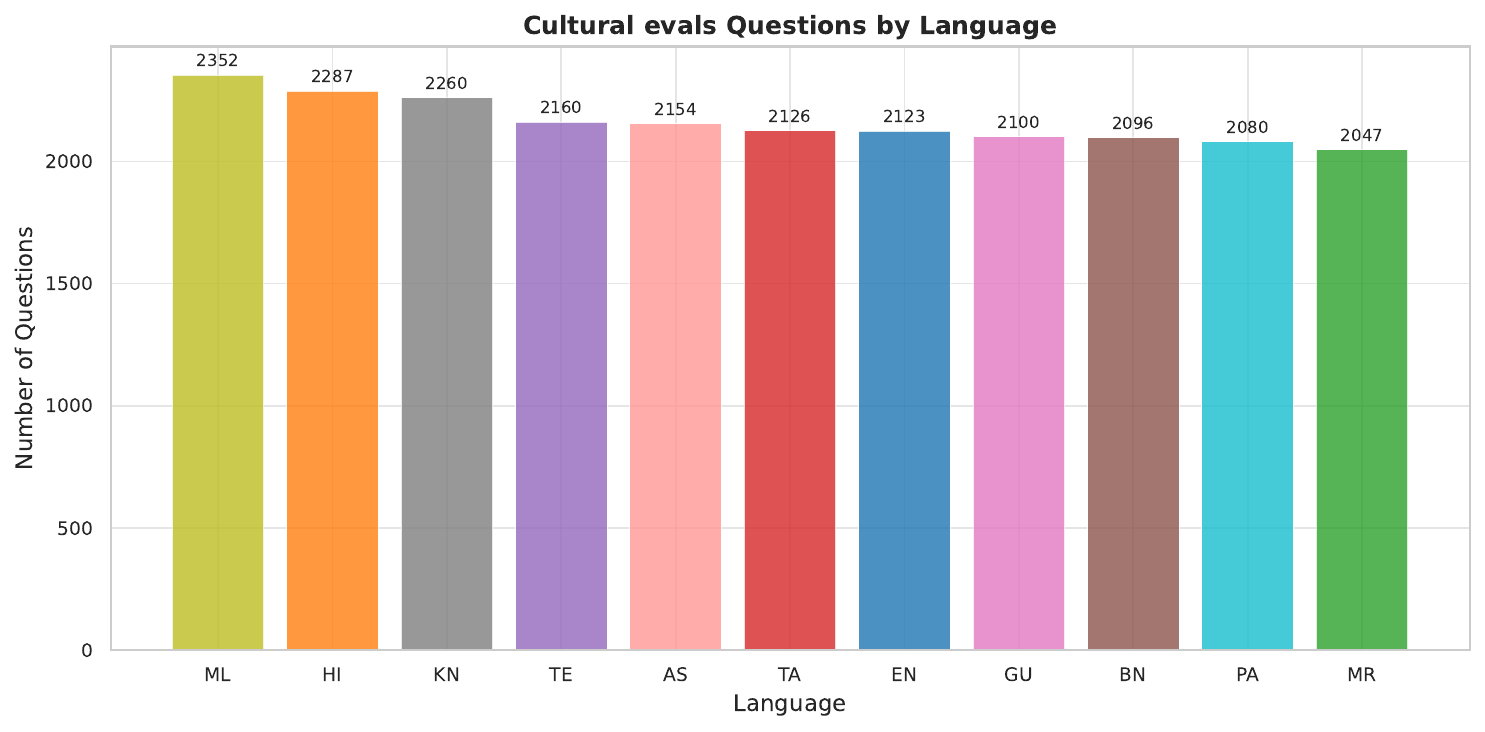}
    \caption{Cultural evaluation language wise distribution}
    \label{fig:cult_evals_lang}
\end{figure*}

To assess the local cultural and community-specific knowledge of Large Language Models (LLMs) within the Indian context, we collected a set of questions in collaboration with a third-party non-profit organization. The dataset consists of 4,399 unique queries that reflect authentic information-seeking behaviors of local populations. These queries span 11 Indian languages (see Figure~\ref{fig:cult_evals_lang}) and cover four high-impact domains: Healthcare, Education, Finance, and Legal. To ensure a robust assessment, we performed a comparative evaluation of all model checkpoints, totaling 91,982 pairwise comparisons (battles).
We refer the reader to \citet{hamna2025buildingbenchmarksgroundup} for the collection process, which we replicated.

\paragraph{Domain Themes.}
The dataset covers a wide spectrum of community needs and themes, some of which were:
\begin{itemize}
    \item \textbf{Education:} Teaching and learning support, and career guidance.
    \item \textbf{Finance:} Insurance, savings, and budgeting strategies.
    \item \textbf{Healthcare:} Senior care protocols and general wellness habits.
    \item \textbf{Legal:} Product/service disputes and family or marriage-related inquiries.
\end{itemize}

\begin{figure*}[!htbp]
    \centering
    \includegraphics[width=0.90\linewidth]{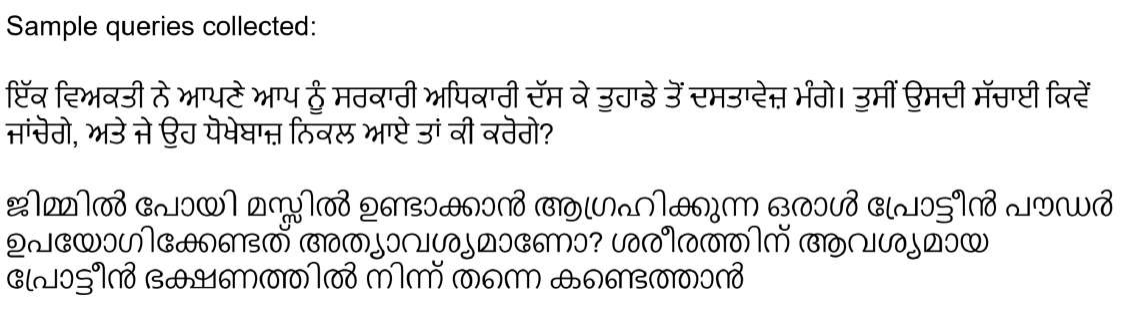}
    \caption{Exanples of a Punjabi and Malayalam query we collected}
    \label{fig:samiksha_sample}
\end{figure*}

\clearpage
\subsection{Sample Prompts Used for Evaluation}\label{sec:sample-prompts}
The following are representative prompts (one per task) used to evaluate different task types. Each prompt follows the established rubrics from Section~\ref{tab:rubrics}.

\subsubsection{Creative Writing - Instruction Adherence}
\label{sec:creative-writing}
    
\begin{PromptBox}{Creative Writing Instruction Adherence Evaluation Prompt}
\textbf{INSTRUCTION:}

You are an expert literary critic and evaluator, tasked with assessing the degree to which a synthetically generated creative piece adheres to the user's writing prompt. You are required to read the USER QUESTION thoroughly and analyze how well the generated response incorporates the specified narrative elements, stylistic choices, and constraints.

\vspace{1em}
\textbf{USER QUESTION:} \\
\texttt{\{user\_prompt\}}

\vspace{1em}
\textbf{TARGET LANGUAGE:} \\
\texttt{\{tgt\_lang\}}

\vspace{1em}
\textbf{ASSISTANT GENERATED RESPONSE:} \\
\texttt{\{assistant\_output\}}

\vspace{1em}
A response that adheres to the user's creative brief has:
\begin{itemize}
  \item \textbf{Narrative \& Thematic Completeness}: It fully incorporates all requested characters, plot points, settings, and themes. The creative piece feels complete and resolves according to the prompt's guidelines.
  \item \textbf{Stylistic \& Tonal Adherence}: The content's tone, mood, and writing style (e.g., genre conventions, a specific author's voice) directly match the user's request.
  \item \textbf{Format \& Constraint Compliance}: It follows all explicit formatting requirements (e.g., poem, script, short story) and abides by any constraints (e.g., word count, inclusion/exclusion of specific words, use of certain literary devices).
  \item \textbf{Creative Intent Alignment}: It successfully captures the spirit and intended artistic goal of the user's prompt, creating a piece that feels like a faithful realization of the user's idea.
\end{itemize}

\vspace{1em}
Use the following scoring scale:
\begin{itemize}
  \item \textbf{5 – Excellent}:  
  The response masterfully incorporates all creative constraints, including plot, character, tone, style, and format. It not only follows the instructions to the letter but also demonstrates a creative flair that enhances the user's original idea.

  \item \textbf{4 – Good}:  
  The response successfully incorporates most creative instructions. There may be minor deviations in tone or style, or a secondary plot/character element might be slightly underdeveloped, but the core creative vision is clearly and effectively realized.

  \item \textbf{3 – Fair}:  
  The response addresses some of the key creative instructions but neglects or misinterprets others. For instance, it might follow the plot but fail to capture the requested tone, or it might ignore a crucial character trait or constraint.

  \item \textbf{2 – Poor}:  
  The response shows significant deviation from the creative brief. It may follow a single, simple instruction (like the general topic) but disregards crucial constraints like genre, character personality, plot structure, or mood.

  \item \textbf{1 – Unacceptable}:  
  The response completely disregards the creative instructions. The generated text is thematically, structurally, and stylistically unrelated to the user's prompt.
\end{itemize}

\vspace{1em}
Return your evaluation in the following JSON format:
\begin{verbatim}
{
  "score": <integer from 1 to 5>,
  "reason": "<brief explanation>"
}
\end{verbatim}

Do not include markdown, comments, or anything outside the JSON.

\end{PromptBox}

\clearpage

\subsubsection{Multi-turn Dialog - Persona Adherence}

\begin{PromptBox}{Multi-turn Dialog Evaluation Prompt}
\textbf{INSTRUCTION:}

You are an expert evaluator tasked with assessing how well a multi-turn dialog maintains persona consistency. You must analyze the conversation to determine if the assistant consistently embodies the specified character or role throughout the interaction.

\vspace{1em}
\textbf{USER QUESTION:} \\
\texttt{\{user\_prompt\}}

\vspace{1em}
\textbf{TARGET LANGUAGE:} \\
\texttt{\{tgt\_lang\}}

\vspace{1em}
\textbf{ASSISTANT GENERATED RESPONSE:} \\
\texttt{\{assistant\_output\}}

\vspace{1em}
A response that demonstrates strong persona adherence has:
\begin{itemize}
  \item \textbf{Character Consistency}: The assistant maintains the same personality traits, speaking style, and behavioral patterns throughout the conversation.
  \item \textbf{Role-Appropriate Knowledge}: The responses reflect knowledge and expertise appropriate to the specified persona.
  \item \textbf{Consistent Voice}: The tone, vocabulary, and manner of speaking remain true to the character across all turns.
  \item \textbf{Believable Interactions}: The persona feels authentic and natural in the conversational context.
\end{itemize}

\vspace{1em}
Use the following scoring scale:
\begin{itemize}
  \item \textbf{5 – Excellent}: Perfect persona consistency with natural, believable character embodiment throughout all turns.
  \item \textbf{4 – Good}: Strong persona adherence with minor inconsistencies that don't break character immersion.
  \item \textbf{3 – Fair}: Generally maintains persona but has noticeable lapses or inconsistencies in character.
  \item \textbf{2 – Poor}: Significant persona inconsistencies that frequently break character immersion.
  \item \textbf{1 – Unacceptable}: Complete failure to maintain persona or embody the specified character.
\end{itemize}

\vspace{1em}
Return your evaluation in the following JSON format:
\begin{verbatim}
{
  "score": <integer from 1 to 5>,
  "reason": "<brief explanation>"
}
\end{verbatim}

Do not include markdown, comments, or anything outside the JSON.
\end{PromptBox}

\subsubsection{Reasoning Tasks - Answer Adequacy}

\begin{PromptBox}{Reasoning Task Evaluation Prompt}
\textbf{INSTRUCTION:}

You are an expert evaluator specializing in logical reasoning and problem-solving. Your task is to assess whether the generated response provides a correct, complete, and well-reasoned answer to the given question.

\vspace{1em}
\textbf{USER QUESTION:} \\
\texttt{\{user\_prompt\}}

\vspace{1em}
\textbf{TARGET LANGUAGE:} \\
\texttt{\{tgt\_lang\}}

\vspace{1em}
\textbf{ASSISTANT GENERATED RESPONSE:} \\
\texttt{\{assistant\_output\}}

\vspace{1em}
A response with excellent answer adequacy demonstrates:
\begin{itemize}
  \item \textbf{Correctness}: The final answer is factually accurate and logically sound.
  \item \textbf{Completeness}: All aspects of the question are addressed without omitting important elements.
  \item \textbf{Direct Relevance}: The response directly answers what was asked without unnecessary tangents.
  \item \textbf{Clear Reasoning}: The logical steps leading to the conclusion are evident and valid.
\end{itemize}

\vspace{1em}
Use the following scoring scale:
\begin{itemize}
  \item \textbf{5 – Excellent}: Completely correct and comprehensive answer with clear, logical reasoning.
  \item \textbf{4 – Good}: Correct answer with minor gaps in completeness or explanation clarity.
  \item \textbf{3 – Fair}: Generally correct but missing some important aspects or contains minor errors.
  \item \textbf{2 – Poor}: Partially correct but has significant errors or omissions in reasoning or conclusion.
  \item \textbf{1 – Unacceptable}: Incorrect answer or completely fails to address the question asked.
\end{itemize}

\vspace{1em}
Return your evaluation in the following JSON format:
\begin{verbatim}
{
  "score": <integer from 1 to 5>,
  "reason": "<brief explanation>"
}
\end{verbatim}

Do not include markdown, comments, or anything outside the JSON.
\end{PromptBox}

\end{document}